\documentclass{article}
\PassOptionsToPackage{numbers, compress}{natbib}
\usepackage[final]{neurips_data_2021}

\usepackage[utf8]{inputenc} 
\usepackage[T1]{fontenc}    
\usepackage{hyperref}       
\usepackage{url}            
\usepackage{booktabs}       
\usepackage{amsfonts}       
\usepackage{nicefrac}       
\usepackage{microtype}      
\usepackage{xcolor}         

\usepackage{amssymb}
\usepackage{amsmath}
\usepackage{mathtools}
\usepackage{amsfonts}
\usepackage{amsthm,bbm}
\usepackage{url}
\usepackage{color}
\usepackage{listings}
\usepackage[para,online,flushleft]{threeparttable}
\usepackage{enumerate}
\usepackage{multirow}
\newcommand{\argmax}{\mathop{\rm argmax}\limits}
\newcommand{\argmin}{\mathop{\rm argmin}\limits}

\newtheorem{example}{Example}
\usepackage{algorithm}
\usepackage{algpseudocode}

\definecolor{dkgreen}{rgb}{0,0.6,0}
\definecolor{customgray}{rgb}{0.25,0.25,0.25}
\definecolor{customred}{rgb}{0.8,0.05,0.05}
\definecolor{customblue}{rgb}{0.05,0.05,0.8}
\usepackage{bbding}
\newcommand{\tick}{\textcolor{dkgreen}{\CheckmarkBold}}
\newcommand{\fail}{\textcolor{red}{\XSolidBrush}}
\newcommand{\best}[1]{\textcolor{customred}{$\textbf{#1}$}}

\newcommand{\worst}[1]{\textcolor{customblue}{$\textbf{#1}$}}
\newcommand{\std}[2]{\large{#1} \textcolor{customgray}{\normalsize{$\pm$#2}}}

\hypersetup{colorlinks=true}
\hypersetup{citecolor=blue}
\hypersetup{linkcolor=blue}

\definecolor{dkred}{rgb}{0.8,0,0}
\definecolor{dkgreen}{rgb}{0,0.4,0}
\definecolor{gray}{rgb}{0.2,0.2,0.2}
\definecolor{mauve}{rgb}{0.7,0,0.9}
\newcommand{\mE}{\mathbb{E}}

\newcommand{\calD}{\mathcal{D}}
\newcommand{\calX}{\mathcal{X}}
\newcommand{\calA}{\mathcal{A}}

\newcommand{\trueV}{V(\pi_e)}
\newcommand{\dm}{\hat{V}_{\mathrm{DM}} (\pi_e; \calD, \hat{q})}
\newcommand{\ipw}{\hat{V}_{\mathrm{IPW}} (\pi_e; \calD)}
\newcommand{\snipw}{\hat{V}_{\mathrm{SNIPW}} (\pi_e; \calD)}

\newcommand{\dr}{\hat{V}_{\mathrm{DR}} (\pi_e; \calD, \hat{q})}
\newcommand{\sndr}{\hat{V}_{\mathrm{SNDR}} (\pi_e; \calD)}
\newcommand{\drs}{\hat{V}_{\mathrm{DRos}} (\pi_e; \calD, \hat{q}, \lambda)}

\newcommand{\switchdr}{\hat{V}_{\mathrm{SwitchDR}} (\pi_e; \calD, \hat{q}, \tau)}

\newcommand{\onpolicy}{V_{\mathrm{on}} (\pi_e)}           

\title{
Open Bandit Dataset and Pipeline:
Towards Realistic and Reproducible Off-Policy Evaluation
}

\author{
    Yuta Saito\thanks{This work was done when YS was at Hanjuku-kaso Co Ltd, Tokyo, Japan.}\\
    Cornell University\\
    \texttt{ys552@cornell.edu}\\
    \And
    Shunsuke Aihara\\
    ZOZO Research\\
    \texttt{shunsuke.aihara@zozo.com}\\
    \AND
    Megumi Matsutani\\
    ZOZO Research\\
    \texttt{megumi.matsutani@zozo.com}\\
    \And
    Yusuke Narita\\
    Yale University\\
    \texttt{yusuke.narita@yale.edu}
}

\begin{document}

\maketitle

\begin{abstract}
  \textit{Off-policy evaluation} (OPE) aims to estimate the performance of hypothetical policies using data generated by a different policy. Because of its huge potential impact in practice, there has been growing research interest in this field. There is, however, no real-world public dataset that enables the evaluation of OPE, making its experimental studies unrealistic and irreproducible. With the goal of enabling realistic and reproducible OPE research, we present \textit{Open Bandit Dataset}, a public logged bandit dataset collected on a large-scale fashion e-commerce platform, ZOZOTOWN. Our dataset is unique in that it contains a set of \textit{multiple} logged bandit datasets collected by running different policies on the same platform. This enables experimental comparisons of different OPE estimators for the first time. We also develop Python software called \textit{Open Bandit Pipeline} to streamline and standardize the implementation of batch bandit algorithms and OPE. Our open data and software will contribute to fair and transparent OPE research and help the community identify fruitful research directions. We provide extensive benchmark experiments of existing OPE estimators using our dataset and software. The results open up essential challenges and new avenues for future OPE research.
\end{abstract}

\section{Introduction}
Interactive bandit systems such as personalized medicine and recommendation platforms produce log data valuable for evaluating and redesigning the system. 
For example, the logs of a news recommendation system records which news article was presented and whether the user read it, giving the system designer a chance to make its recommendations more relevant. 
Exploiting log bandit data is, however, more difficult than conventional supervised machine learning, as the result is only observed for the action chosen by the system, but not for all the other actions that the system could have taken. 
The logs are also biased in that they overrepresent the actions favored by the system. 
A potential solution to this problem is an A/B test, which compares the performance of counterfactual systems in an online environment. 
However, A/B testing counterfactual systems is often difficult because deploying a new policy is time- and money-consuming and entails the risk of failure.
This leads us to the problem of \textit{off-policy evaluation} (OPE), which aims to estimate the performance of a counterfactual (or evaluation) policy using only log data collected by a past (or behavior) policy.
OPE allows us to evaluate the performance of candidate policies without implementing A/B tests and contributes to safe policy improvement.
Its applications range from contextual bandits~\citep{bottou2013counterfactual,li2010contextual,li2011unbiased,narita2019efficient,saito2021counterfactual,strehl2010learning, swaminathan2015batch,swaminathan2015self, swaminathan2017off,uehara2020off,wang2016optimal} and reinforcement learning in the web industry~\citep{farajtabar2018more,jiang2016doubly,kallus2019intrinsically,liu2018representation,thomas2016data,thomas2015confidence,xie2019towards} to other social domains such as healthcare~\citep{murphy2001marginal} and education~\citep{mandel2014offline}.

\paragraph{Issues with current experimental procedures.}
Although the research community has produced theoretical breakthroughs over the past decade, the experimental evaluation of OPE remains primitive. 
Specifically, it lacks a public benchmark dataset for comparing the performance of different estimators. 
Researchers often validate their estimators using synthetic simulation environments~\citep{kallus2019intrinsically,liu2018representation,uehara2020off,voloshin2019empirical,xie2019towards}. 
A version of the synthetic approach is to modify multiclass classification datasets and treat supervised machine learning methods as bandit policies to evaluate the estimation accuracy of OPE estimators~\citep{dudik2014doubly,farajtabar2018more,vlassis2019design,wang2016optimal}. 
An obvious problem with these studies is that they are \textit{\textbf{unrealistic}} because there is no guarantee that their simulation environment is similar to real-world settings. 
To solve this issue, some previous studies use proprietary real-world datasets~\citep{gilotte2018offline,gruson2019offline,narita2019efficient,narita2020safe}.
Because these datasets are not public, however, the results are \textit{\textbf{irreproducible}}, and it remains challenging to compare existing estimators with new ideas in a fair manner. 
The lack of a public real-world benchmark makes it hard to identify critical research challenges and the bottleneck of the literature.
This contrasts with other domains of machine learning, where large-scale open datasets, such as the ImageNet dataset~\citep{deng2009imagenet}, have been pivotal in driving objective progress~\citep{dwivedi2020benchmarking,girshick2014rich,he2016deep,hu2020open}.

\paragraph{Contributions.}
Our goal is to implement and evaluate OPE in \textit{\textbf{realistic and reproducible}} ways. 
To this end, we release \textit{Open Bandit Dataset}, a set of logged bandit datasets collected on the ZOZOTOWN platform.\footnote{https://corp.zozo.com/en/service/}
ZOZOTOWN is the largest fashion e-commerce platform in Japan.
When the platform produced the data, it used Bernoulli Thompson Sampling (Bernoulli TS) and uniform random (Random) policies to recommend fashion items to users.
The dataset thus includes a set of \textit{two} logged bandit datasets collected during an A/B test of these bandit policies.
Having multiple (at least two) log datasets is essential because it enables the evaluation of the estimation accuracy of OPE estimators as we describe in detail in Section~\ref{sec:benchmark_experiment}.

In addition to the dataset, we implement \textit{Open Bandit Pipeline}, an open-source Python software including a series of modules for implementing dataset preprocessing, policy learning methods, and OPE estimators.
Our software provides a complete, standardized experimental procedure for OPE research, ensuring that performance comparisons are fair, transparent, and reproducible.
It also enables fast and accurate OPE implementation through a single unified interface, simplifying the practical use of OPE.

Using our dataset and software, we perform comprehensive benchmark experiments on existing estimators.
We implement this OPE experiment by using the log data of one of the policies (e.g., Bernoulli TS) to estimate the policy value of the other policy (e.g., Random) with each OPE estimator. 
We then assess the accuracy of the estimator by comparing its estimation with the policy value obtained from the data in an \textit{on-policy} manner.
Through the experiments, we showcase the utility of Open Bandit Dataset and Pipeline by using them to analyze the challenges that we face when we try applying OPE to real-world scenarios.

Our key contributions are summarized as follows:
\begin{itemize}
    \item \textbf{Public Dataset}: We build and release \textit{Open Bandit Dataset}, a set of \textit{two} logged bandit data to enable realistic and reproducible research on OPE.
    \item \textbf{Software Implementation}: We implement \textit{Open Bandit Pipeline}, an open-source Python software that helps practitioners utilize OPE to evaluate their bandit systems. It also helps researchers compare different OPE estimators in a standardized manner.
    \item \textbf{Benchmark Experiment}: We perform comprehensive benchmark experiments on existing OPE estimators and indicate critical challenges in future research.
\end{itemize}

\section{Off-Policy Evaluation} \label{sec:ope}

\subsection{Setup}\label{sec:setup}
We consider a general contextual bandit setting.
Let $r \in [0, r_{\mathrm{max}}]$ denote a reward variable (e.g., whether a fashion item as an action results in a click). 
We let $x \in \calX$ be a context vector (e.g., the user's demographic profile) that the decision maker observes when picking an action. 
We also let  $a \in \calA$ be a discrete action such as a fashion item in a recommender system. 
Rewards and contexts are sampled from unknown distributions $p (r \mid x, a)$ and $p(x)$, respectively.
We call a function $\pi: \calX \rightarrow \Delta(\calA)$ a \textit{policy}.
It maps each context $x \in \calX$ into a distribution over actions, where $\pi (a \mid x)$ is the probability of taking action $a$ given context $x$.
We describe some examples of such decision making policies in Appendix~\ref{app:examples}.

Let $\calD := \{(x_i,a_i,r_i)\}_{i=1}^n$ be logged bandit dataset with $n$ observations.
$a_i$ is a discrete variable indicating which action in $\mathcal{A}$ is chosen for $i$.
$r_i$ and $x_i$ denote the reward and the context observed for each data, respectively.
We assume that the logged dataset is generated by \textit{behavior policy} $\pi_b$ as 
$$\{(x_i,a_i,r_i)\}_{i=1}^n \ \sim \ \prod_{i=1}^n p(x_i) \pi_b (a_i \mid x_i) p(r_i \mid x_i, a_i),$$
where each triplet is sampled independently from the product distribution.
We sometimes use $\mE_{n} [f] := n^{-1} \sum_{(x_i, a_i, r_i) \in \calD} f(x_i, a_i, r_i)$ to denote the empirical expectation over $n$ observations in $\calD$.
We also use $q(x,a) := \mE_{r \sim p(r|x,a)} [ r \mid x, a ]$ and $g(x,\pi) := \mE_{a \sim \pi(a|x)}[g(x,a) \mid x] $ to define estimators.

\subsection{Estimation Target and Estimators}
We are interested in using logged bandit data to estimate the following \textit{policy value} of any given \textit{evaluation policy} $\pi_e$, which might be different from $\pi_b$: 
\begin{align*}
    \trueV := \mE_{(x,a,r) \sim p(x) \pi_e (a \mid x) p(r \mid x, a)} [r].
\end{align*}

An OPE estimator $\hat{V}$ estimates $\trueV$ using only $\calD$ as $\trueV \approx \hat{V} (\pi_e; \mathcal{D})$.
We define three standard estimators in the following.\footnote{We define some other advanced estimators in Appendix~\ref{app:estimators}.} 

\paragraph{Direct Method (DM).}
DM~\citep{beygelzimer2009offset} first estimates $q$ using a supervised machine learning model, such as random forest or ridge regression. 
It then plugs it in to estimate the policy value as
\begin{align*}
    \dm := \mE_{n} [ \hat{q} (x_i, \pi_e) ], 
\end{align*}
where $\hat{q}(x,a)$ is a reward estimator.
If $\hat{q}(x,a)$ is accurate, DM also estimates the policy value accurately. 
If $\hat{q}(x,a)$ is inaccurate, however, the final estimator is no longer consistent.
The model misspecification issue is problematic because the extent of misspecification cannot be easily quantified from data~\citep{farajtabar2018more,voloshin2019empirical}.

\paragraph{Inverse Probability Weighting (IPW).}
To alleviate the issue with DM, IPW is often used~\citep{precup2000eligibility,strehl2010learning}. 
This estimator weighs the observed rewards by the importance weights as
\begin{align*}
    \ipw := \mE_{n} [w(x_i,a_i) r_i ],
\end{align*}
where $w(x,a) := \pi_e(a \mid x) / \pi_b(a \mid x)$.
When the behavior policy is known, IPW is unbiased and consistent.
However, it can have a large variance, especially when the evaluation policy deviates significantly from the behavior policy.

\paragraph{Doubly Robust (DR).}
DR~\citep{dudik2014doubly} combines DM and IPW as follows.
\begin{align*}
    \dr := \mE_{n} [ \hat{q} (x_i, \pi_e) + w(x_i,a_i)  (r_i-\hat{q}(x_i, a_i) ) ].
\end{align*}
DR mimics IPW to use a weighted version of rewards, but it also uses $\hat{q}$ as a control variate to decrease the variance. 
It preserves the consistency of IPW if either the importance weight or the reward estimator is consistent (a property called \textit{double robustness}).
Moreover, DR is \textit{semiparametric efficient} when the reward estimator is correctly specified~\citep{narita2019efficient}.
However, when it is misspecified, this estimator can have a larger asymptotic mean-squared-error than that of IPW~\citep{kallus2019intrinsically}.

\begin{figure*}[ht]
    \centering
    \includegraphics[clip, width=7cm]{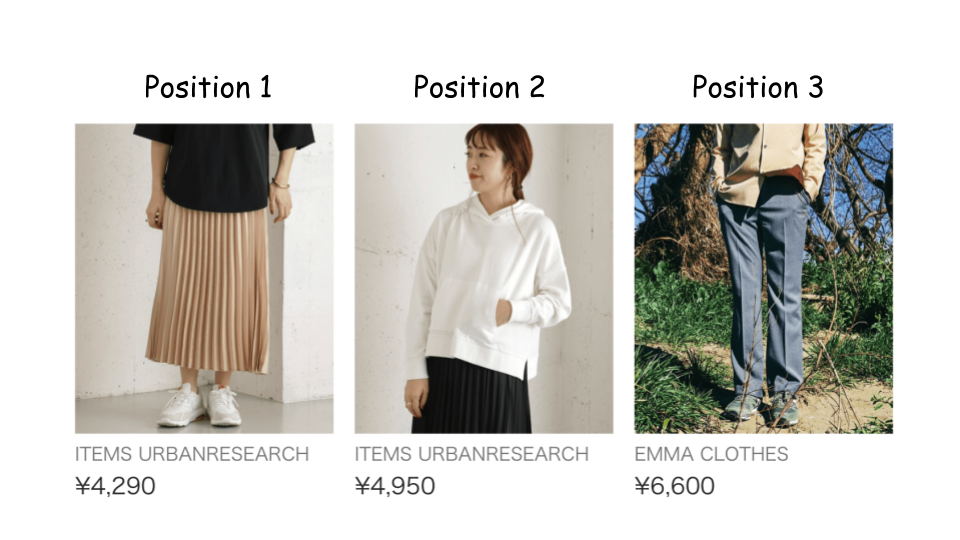}
    \caption{Fashion items as actions displayed in ZOZOTOWN recommendation interface.}
    \label{fig:displayed_fashion_item_sample}
\end{figure*}

\begin{table*}[ht]
\large
\centering
\caption{Statistics of Open Bandit Dataset}
\def\arraystretch{1.2}
\scalebox{0.68}{
\begin{tabular}{cccccccc}
\toprule
\textbf{Campaigns} & \textbf{Data Collection Policies} && \textbf{\#Data} & \textbf{\#Items} & \textbf{\#Dim} & \textbf{\std{CTR ($V (\pi)$)}{95\% CI}} & \textbf{Relative-CTR} \\
\midrule \midrule
\multirow{2}{*}{\textbf{ALL}} 
& \textbf{Random} && 1,374,327 & \multirow{2}{*}{80} & \multirow{2}{*}{84} & \std{0.35\%}{0.010} & 1.00 \\
& \textbf{Bernoulli TS} && 12,168,084 &  &  & \std{0.50\%}{0.004} & 1.43 \\ 
\midrule
\multirow{2}{*}{\textbf{Men's}} 
& \textbf{Random} && 452,949 & \multirow{2}{*}{34} & \multirow{2}{*}{38} & \std{0.51\%}{0.021} & 1.48 \\
& \textbf{Bernoulli TS} && 4,077,727 &  &  & \std{0.67\%}{0.008} & 1.94 \\ 
\midrule
\multirow{2}{*}{\textbf{Women's}}
& \textbf{Random} && 864,585 & \multirow{2}{*}{46} & \multirow{2}{*}{50} & \std{0.48\%}{0.014} & 1.39 \\
& \textbf{Bernoulli TS} && 7,765,497 &  &  & \std{0.64\%}{0.056} & 1.84 \\ 
\bottomrule
\end{tabular}
}
\vskip 0.05in
\raggedright
\fontsize{9.0pt}{9.0pt}\selectfont \textit{Note}: 
Bernoulli TS stands for Bernoulli Thompson Sampling. 
\textbf{\#Data} is the total number of user impressions observed during the 7-day experiment. 
\textbf{\#Items} is the total number of items having a non-zero probability of being recommended by each policy.
\textbf{\#Dim} is the number of dimensions of the raw context vectors.
\textbf{CTR} is the percentage of a click being observed in the log data, and this is the performance of the data collection policies in each campaign.
95\% confidence interval (CI) of CTR is calculated based on the normal approximation of the Bernoulli sampling.
\textbf{Relative-CTR} is CTR relative to that of the Random policy for the ``ALL'' campaign.
\label{tab:obp_stats}
\end{table*}

\section{Open-Source Dataset and Software} \label{sec:dataset}
Motivated by the paucity of real-world datasets and implementations enabling the evaluation of OPE, we release the following open-source dataset and software.

\paragraph{Open Bandit Dataset.}
Our open-source dataset is a set of \textit{two} logged bandit datasets provided by ZOZO, Inc., the largest fashion e-commerce company in Japan.
The company uses multi-armed bandit algorithms to recommend fashion items to users in their fashion e-commerce platform called ZOZOTOWN.
We present examples of displayed fashion items in Figure~\ref{fig:displayed_fashion_item_sample}.
We collected the data in a 7-day experiment in late November 2019 on three “campaigns,'' corresponding to ``ALL'', ``Men's'', and ``Women's'' items, respectively.
Each campaign randomly uses either the Random policy or the Bernoulli TS policy for each user impression.
These policies select three of the candidate fashion items for each user.
Figure~\ref{fig:displayed_fashion_item_sample} shows that there are three \textit{positions} in our dataset.
We assume that the reward (click indicator) depends only on the item and its position, which is a general assumption on the click generative process used in the web industry~\citep{li2018offline}.
Under this assumption, we can apply the OPE setup in Section~\ref{sec:ope} to our dataset.
We provide some statistics of the dataset in Table~\ref{tab:obp_stats}.
The dataset is large and contains many millions of recommendation instances. 
Each row of the data has feature vectors such as age, gender, and past click history of the users. 
These feature vectors are hashed, thus the dataset does not contain any personally identifiable information. 
Moreover, the dataset includes some item-related features such as price, fashion brand, and item categories. 
It also includes the probability that item $a$ is displayed at each position by the data collection policies.
This probability is used to calculate the importance weights.\footnote{We computed the action choice probabilities by Monte Carlo simulations based on the policy parameters (e.g., parameters of the beta distribution used by Bernoulli TS) used during the data collection process.} 
We share the full version of our dataset at \url{https://research.zozo.com/data.html}.\footnote{The dataset is licensed under CC BY 4.0.}
Small-sized example data are also available at \url{https://github.com/st-tech/zr-obp/tree/master/obd}.

To our knowledge, our open-source dataset is the first to include logged bandit datasets collected by running \textit{multiple} (at least two) different policies and the exact policy implementations used in real production, enabling \textbf{\textit{realistic and reproducible evaluation of OPE}} for the first time.
Indeed, Open Bandit Dataset is already actively used by multiple research papers to benchmark new bandit algorithms and OPE estimators~\citep{dai2021offline,kato2020practical,komiyama2021finite,saito2021evaluating,tanimoto2021causal}.

\paragraph{Open Bandit Pipeline.}
To facilitate the use of OPE in practice and standardize its experimental procedures, we implement a Python software called \textit{Open Bandit Pipeline}.\footnote{https://github.com/st-tech/zr-obp}
Our software contains the following main modules:
\begin{itemize}
    \item The \textbf{dataset} module provides a data loader to preprocess Open Bandit Dataset and tools to generate synthetic bandit datasets. It also implements functions to handle multiclass classification datasets as bandit data, which is useful when we conduct OPE experiments in research papers.
    \item The \textbf{policy} module implements several online bandit algorithms and off-policy learning methods such as a neural network-based off-policy learning method.
    This module also implements interfaces that allow practitioners to easily evaluate their own policies in their business using OPE.
    \item The \textbf{ope} module implements several existing OPE estimators including DM, IPW, DR, and some advanced ones such as Switch~\citep{wang2016optimal}, More Robust Doubly Robust (MRDR)~\citep{farajtabar2018more}, and DR with Optimistic Shrinkage (DRos)~\citep{su2020doubly}.
    This module also provides generic abstract interfaces to support custom implementations so that researchers can evaluate their own estimators easily.
\end{itemize}
Appendix~\ref{app:obp_usage} describes how the software facilitates the evaluation of OPE and bandit algorithms.
We also prepare plenty of tutorial contents at \url{https://github.com/st-tech/zr-obp/tree/master/examples/quickstart} to help users grasp the usage of the software easily.
We also provide the thorough documentation at \url{https://zr-obp.readthedocs.io/en/latest/}.

Every core function of the software is tested and thus are well maintained.\footnote{https://github.com/st-tech/zr-obp/tree/master/tests}
It currently has five core contributors.\footnote{https://github.com/st-tech/zr-obp/graphs/contributors}
The active development and maintenance will continue over a long period.
Users can follow our progress at the following mailing list: \url{https://groups.google.com/g/open-bandit-project}.

We believe that the software allows researchers to focus on building their OPE estimator and to easily compare it with other methods in a standardized manner. 
It will also help practitioners implement cutting-edge estimators in their applications and improve their decision making systems.

\begin{table}[t]
\large
\centering
\caption{Comparison of currently available bandit datasets} \label{tab:dataset}
\def\arraystretch{1.2}
\scalebox{0.625}{
\begin{tabular}{cccc}
\toprule
& \textbf{Criteo Data}~\citep{lefortier2016large} & \textbf{Yahoo! R6A\&B}~\citep{li2010contextual} & \textbf{Open Bandit Dataset} (ours) \\
\midrule \midrule
\textbf{Domain} & Display Advertising & News Recommendation & Fashion E-Commerce \\
\textbf{Dataset Size} &  $\approx$ 103M  & $\approx$ 40M & $\approx$ 26M  \\
\textbf{\#Data Collection Policies} & 1 & 1 & \textbf{2} \\
\textbf{Uniform Random Data} & \fail & \tick  & \tick \\
\textbf{Data Collection Policy Code} & \fail & \fail  & \tick \\
\textbf{Evaluation of Bandit Algorithms} & \tick & \tick  & \tick \\
\textbf{Evaluation of OPE} & \fail & \fail  & \tick \\
\textbf{Pipeline Implementation} & \fail & \fail & \tick \\
\bottomrule
\end{tabular}
}
\vskip 0.1in
\raggedright
\fontsize{9.0pt}{9.0pt}\selectfont \textit{Note}: 
\textbf{Dataset Size} is the total number of samples included in the whole dataset.
\textbf{\#Data Collection Policies} is the number of policies that were used to collect the data.
\textbf{Uniform Random Data} indicates whether the dataset contains a subset of data generated by the uniform random policy.
\textbf{Data Collection Policy Code} indicates whether the code to replicate data collection policies is publicized.
\textbf{Evaluation of Bandit Algorithms} indicates whether it is possible to use the data to evaluate bandit algorithms.
\textbf{Evaluation of OPE} indicates whether it is possible to use the dataset to evaluate OPE estimators.
\textbf{Pipeline Implementation} indicates whether a pipeline tool to handle the dataset is available.
\vskip 0.1in
\large
\centering
\caption{Comparison of currently available packages of bandit algorithms and OPE} \label{tab:package}
\def\arraystretch{1.2}
\scalebox{0.6}{
\begin{tabular}{ccccc}
\toprule
& \textbf{Vowpal Wabbit}~\citep{bietti2018contextual} & \textbf{contextualbandits}~\citep{cortes2018adapting} & \textbf{RecoGym}~\citep{rohde2018recogym} & \textbf{Open Bandit Pipeline} (ours) \\
\midrule \midrule
\textbf{Synthetic Data Generator} & \fail & \fail  & \tick & \tick  \\
\textbf{Classification Data Handler} & \fail & \fail & \fail & \tick \\
\textbf{Support for Real-World Data} & \fail & \fail & \fail & \tick \\
\textbf{Bandit Algorithms} & \tick & \tick & \tick  & \tick \\
\textbf{Basic OPE Estimators} & \tick & \tick & \fail & \tick \\
\textbf{Advanced OPE Estimators} & \fail & \fail & \fail  & \tick \\
\textbf{Evaluation of OPE} & \fail & \fail & \fail  & \tick \\
\bottomrule
\end{tabular}
}
\vskip 0.1in
\raggedright
\fontsize{9.0pt}{9.0pt}\selectfont \textit{Note}: 
\textbf{Synthetic Data Generator} indicates whether it is possible to generate synthetic bandit data with the package.
\textbf{Classification Data Handler} indicates whether it is possible to transform multiclass classification data to bandit data with the package.
\textbf{Support for Real-World Data} indicates whether it is possible to handle real-world bandit data with the package.
\textbf{Bandit Algorithms} indicates whether the package includes implementations of online and offline bandit algorithms.
\textbf{Basic OPE Estimators} indicates whether the package includes implementations of \textit{basic} OPE estimators such as DM, IPW, and DR described in Section~\ref{sec:ope}.
\textbf{Advanced OPE Estimators} indicates whether the package includes implementations of \textit{advanced} OPE estimators such as Switch and More Robust Doubly Robust described in Appendix~\ref{app:estimators}.
\textbf{Evaluation of OPE} indicates whether it is possible to evaluate the accuracy of OPE estimators with the package.
\end{table}

\section{Related Work} \label{sec:related}
Here, we summarize the existing related work and resources, and clarify the advantages of ours.

\paragraph{Related Datasets.}
Our dataset is closely related to those of~\citep{lefortier2016large} and~\citep{li2010contextual}. 
Lefortier et al.~\citep{lefortier2016large} introduce a large-scale logged bandit data (Criteo Data\footnote{\href{https://www.cs.cornell.edu/~adith/Criteo/}{https://www.cs.cornell.edu/\~adith/Criteo/}}) from a leading company in display advertising, Criteo.
The dataset contains context vectors of user impressions, advertisements (ads) as actions, and click indicators as rewards.
It also provides the ex-ante probability of each ad being selected by the behavior policy.
Therefore, this dataset can be used to compare different off-policy \textit{learning} methods, which aim to learn a new policy using only logged bandit data.
In contrast, Li et al.~\citep{li2010contextual} introduce a dataset (Yahoo! R6A\&B\footnote{\url{https://webscope.sandbox.yahoo.com/catalog.php?datatype=r}}) collected on a news recommendation interface of the Yahoo! Today Module.
The dataset contains context vectors of user impressions, presented news as actions, and click indicators as rewards.
The dataset was collected by running a uniform random policy on the news recommendation platform, allowing researchers to evaluate their (online or offline) bandit algorithms.

However, the existing datasets have several limitations, which we overcome as follows:
\begin{itemize}
    \item The existing datasets include only a single logged bandit dataset collected by running only a single policy.
    Moreover, the previous datasets do not provide the implementation to replicate the policies used during data collection.
    As a result, these datasets cannot be used for the comparison of different OPE estimators, although they can be used to evaluate policy \textit{learning} methods. 
    
    $\rightarrow$ In contrast, we provide the code to replicate the data collection policies in our software, which allows researchers to rerun the same policies on the log data. Without the code of the exact algorithms, we could not implement the evaluation of OPE experiments. Therefore, the code and algorithm release is an essential component of our open-source.
    Moreover, our dataset consists of a set of \textit{two} different logged bandit datasets generated by running two different policies on the same platform. 
    It enables the  comparison of different OPE estimators, as we show in Section~\ref{sec:benchmark_experiment}.
    
    \item The existing datasets do not provide a tool to handle their data.
    Researchers have to reimplement the experimental environment by themselves before implementing their own OPE estimators.
    This can lead to inconsistent experimental conditions across different studies, potentially causing reproducibility issues.
    
    $\rightarrow$ We implement Open Bandit Pipeline to simplify and standardize the experimental processing of bandit algorithms and OPE. This tool thus contributes to the reproducible and transparent use of our dataset.
\end{itemize}

Table~\ref{tab:dataset} summarizes the key differences between our dataset and the existing ones.

\paragraph{Related Packages.}
There are several existing packages related to Open Bandit Pipeline.
\textit{Vowpal Wabbit}\footnote{https://github.com/VowpalWabbit/vowpal\_wabbit} is a library for fast machine learning, online learning, contextual bandits, and reinforcement learning~\citep{bietti2018contextual}.
It handles learning problems with any number of sparse features, achieving great scaling.
The \textit{contextualbandits} package\footnote{https://github.com/david-cortes/contextualbandits} contains implementations of several contextual bandit algorithms~\citep{cortes2018adapting}. 
It aims to provide an easy procedure to compare bandit algorithms to reproduce research papers that do not provide easily available implementations.
There is also \textit{RecoGym}\footnote{https://github.com/criteo-research/reco-gym}, which focuses on providing simulation bandit environments imitating the e-commerce recommendation setting~\citep{rohde2018recogym}.

However, the following features differentiate our software from the previous ones:
\begin{itemize}
    \item The previous packages focus on implementing and comparing online bandit algorithms or off-policy learning methods.
    However, they \textit{\textbf{cannot}} be used to implement several advanced OPE estimators. 
    They also do not help conduct benchmark experiments of OPE estimators.
    
    $\rightarrow$ Our software implements a wide variety of OPE estimators, including advanced ones such as Switch, MRDR, and DRos.
    In addition, Open Bandit Pipeline provides estimators that can address continuous actions or combinatorial actions, which cannot be handled by other packages.
    It also provides flexible interfaces for implementing new OPE estimators, allowing researchers to plug in their own estimators and compare them with existing estimators easily.
    
    \item The previous packages accept their own interface and data formats. Thus, they are not user-friendly.
    
    $\rightarrow$ Our software follows the prevalent \textit{scikit-learn} style interface and provides sufficient example codes at \url{https://github.com/st-tech/zr-obp/tree/master/examples} so that anyone, including practitioners and students, can follow the usage.
    
    \item The previous packages cannot handle real-world bandit datasets.
    
    $\rightarrow$ Our software comes with Open Bandit Dataset and includes the \textbf{dataset module}.
    This facilitates the evaluation of bandit algorithms and OPE estimators using real-world data.
\end{itemize}

Table~\ref{tab:package} summarizes the key differences between our software and the existing ones.

\paragraph{Related Benchmarks.}
There are several studies conducting benchmark experiments on OPE estimators.
Fu et al.~\citep{fu2021benchmarks} provide the Deep Off-Policy Evaluation (DOPE) benchmark, which is designed to evaluate the performance of OPE estimators on several control tasks.
The notable contribution of DOPE is that it evaluates the OPE performance across a range of evaluation policies with different policy values and measures performance on ranking and selection as well as policy evaluation.
Evaluating the ranking and policy selection performance is challenging with real-world bandit data due to the difficulty of deploying many policies during data collection. 
Voloshin et al.~\citep{voloshin2019empirical} provide a benchmark study on a variety of tasks ranging from tabular problems to image-based tasks in Atari. 
A wide variety of OPE estimators are compared in the benchmark, including recent variants of DR such as MRDR and Self-Normalized DR (SNDR), producing a holistic summary of the challenges one should address in OPE applications.

Our work differentiates itself from these previous benchmark studies in several ways.
First, our benchmark and implementation cover relevant methods that are not included in the previous benchmarks.
For example, we evaluate some advanced estimators such as Switch Doubly Robust (Switch-DR), and DRos, which are not compared in DOPE.
Moreover, we evaluate how a recent hyperparameter tuning method proposed in~\citep{su2020doubly} works with real-world bandit data.
Tuning hyperparameters of OPE estimators is a critical component in OPE application, as it can greatly affect the OPE performance.
However, the previous benchmarks do not evaluate the performance of the tuning method.
Moreover, we release the real-world bandit dataset that allows benchmarking of OPE estimators in a realistic scenario.
Our public dataset makes it possible to identify what matters in applying OPE to real-world scenarios.
This is in contrast to the previous benchmarks using only synthetic environments.
Indeed, we found some critical bottlenecks, which have not yet been pointed out in the literature. Specifically, we found that it is necessary  to develop a reliable method to choose and tune OPE estimators in a data-driven manner (discussed in Sections~\ref{sec:benchmark_experiment} and~\ref{sec:conclusion}). We would also emphasize that the previous benchmarks focus on conducting comprehensive empirical studies. They do not provide implementations that allow researchers to add their own estimators to the benchmark, generate synthetic data, and handling real bandit data.

\section{Benchmark Experiments} \label{sec:benchmark_experiment}
We perform benchmark experiments of OPE estimators using Open Bandit Dataset and Pipeline.
We first describe an experimental protocol to evaluate OPE estimators and use it to compare a wide variety of existing estimators.
We then discuss the initial findings from the experiments.
We share the code to replicate the benchmark experiments at \url{https://github.com/st-tech/zr-obp/tree/master/benchmark/ope}.

\subsection{Experimental Protocol}
We can empirically evaluate OPE estimators' performance by using two sources of logged bandit data collected by running two different policies.
In the protocol, we regard one policy as behavior policy $\pi_b$ and the other as evaluation policy $\pi_e$.
We denote log data generated by $\pi_b$ and $\pi_e$ as $\calD^{(b)} := \{ (x^{(b)}_i, a^{(b)}_i, r^{(b)}_i) \}_{i=1}^{n^{(b)}}$ and $\calD^{(e)} := \{ (x^{(e)}_i, a^{(e)}_i, r^{(e)}_i) \}_{i=1}^{n^{(e)}}$.
Then, by applying the following protocol to several different OPE estimators, we compare their estimation performance:
\begin{enumerate}
    \item Estimate the policy value of $\pi_e$ using $\calD^{(b)}$ by OPE estimator $\hat{V}$. We represent a policy value estimated by $\hat{V}$ as $\hat{V} (\pi_e; \calD^{(b)})$.
    \item Evaluate the estimation accuracy of $\hat{V}$ using the following \textit{squared error} (SE):
    \begin{align*}
        \textit{SE} (\hat{V}; \calD^{(b)}) := \left( \hat{V}(\pi_e; \calD^{(b)}) -  \onpolicy \right)^2,
    \end{align*}
    where $\onpolicy := (1/n^{(e)}) \sum_{i=1}^{n^{(e)}} r^{(e)}_i $ is the Monte-Carlo estimate (on-policy estimate) of $V(\pi_e)$ based on $\calD^{(e)}$. 
    \item Repeat the above process $T$ times with different bootstrap samples and calculate the following \textit{root mean-squared-error} (RMSE) as the estimators' performance measure.
    \begin{align*}
        \textit{RMSE} (\hat{V}; \calD^{(b)}) & := \sqrt{\frac{1}{T} \sum_{t=1}^T \textit{SE} (\hat{V}; \calD_t^{(b,*)})},
    \end{align*}
    where $\calD_t^{(b,*)}$ is the $t$-th bootstrapped sample of $\calD^{(b)}$.
\end{enumerate}

Algorithm~\ref{algo:evaluation_of_ope} in Appendix~\ref{app:experimental_settings} describes the experimental protocol to evaluate OPE estimators in detail.

\subsection{Compared Estimators}
We compare the following OPE estimators:
DM, IPW, Self-Normalized Inverse Probability Weighting (SNIPW), DR, SNDR, Switch-DR, and DRos.
We tune the built-in hyperparameter of Switch-DR and DRos using a data-driven hyperparameter tuning method described in Su et al.~\citep{su2020doubly}.
The details of the above estimators and data-driven hyperparameter tuning method are given in Appendix~\ref{app:estimators}.

For estimators except for DM, we use the true action choice probability $\pi_b(a | x)$ contained in Open Bandit Dataset.
For estimators except for IPW and SNIPW, we need to obtain a reward estimator $\hat{q}$. 
We do this by using gradient boosting\footnote{Specifically, we use `sklearn.ensemble.HistGradientBoostingClassifier(learning\_rate=0.01, max\_iter=100, max\_depth=5, min\_samples\_leaf=10, random\_state=12345)' to obtain $\hat{q}$.} (implemented in \textit{scikit-learn}~\citep{scikit-learn}) and training it on $\calD^{(b)}$.
We also use cross-fitting~\citep{kallus2019efficiently,narita2020safe} to avoid substantial bias from overfitting when obtaining $\hat{q}$.

\begin{table*}[t]
\begin{minipage}{1.0\textwidth}
\caption{RMSE ($\times 10^3$) of OPE estimators (\textbf{Bernoulli TS $\rightarrow$ Random})} \label{tab:se_ts}
\vskip 0.05in
\large
\centering
\def\arraystretch{1.25}
\scalebox{0.65}{
\begin{tabular}{c|cccccc}
\toprule
 \textbf{OPE Estimators ($\hat{V}$)} && \textbf{ALL} && \textbf{Men's} && \textbf{Women's} \\
\midrule \midrule
 \textbf{IPW}$^1$ && 0.493 (1.560)$^3$ && 0.789 (1.719) && 0.776 (1.382)$^{3}$ \\
 \textbf{SNIPW}$^2$ && 0.507 (1.602)$^3$ && 0.644 (1.403)$^{1/3}$ && 0.804 (1.433)$^{3}$ \\
 \textbf{DM}$^3$ && \worst{1.026 (3.244)} && 0.773 (1.685) && \worst{0.816 (1.455)} \\
 \textbf{DR}$^4$ && 0.482 (1.526)$^3$ && 0.613 (1.336)$^{1/3}$ && 0.803 (1.430)$^{3}$ \\
 \textbf{SNDR}$^5$ && 0.482 (1.526)$^3$ && 0.659 (1.436)$^{1/3}$ && 0.791 (1.408)$^{3}$ \\
 \textbf{Switch-DR}$^6$ && 0.482 (1.526)$^3$ && 0.613 (1.336)$^{1/3}$ && 0.803 (1.430)$^{3}$ \\
 \textbf{DRos}$^7$ && \best{0.316 (1.000)}$^{1/2/3/4/5/6}$ && \best{0.459 (1.000)}$^{1/2/3/4/5/6}$ && \best{0.561 (1.000)}$^{1/2/3/4/5/6}$ \\
\bottomrule
\end{tabular}}
\end{minipage}
\\
\vspace{0.1in}
\\
\begin{minipage}{1.\textwidth}
\caption{RMSE ($\times 10^3$) of DRos with different hyperparameter values (\textbf{ALL Campaign})}  \label{tab:hyperparam}
\vspace{0.05in}
\large
\centering
\def\arraystretch{1.2}
\scalebox{0.625}{
\begin{tabular} {c|ccc}
\toprule
\textbf{$\lambda$ of DRos}  && \textbf{Random $\rightarrow$ Bernoulli TS} & \textbf{Bernoulli TS $\rightarrow$ Random} \\ 
\midrule \midrule
\textbf{1} && 1.384 (2.906) & 0.963 (3.920)  \\
\textbf{5} && 1.247 (2.619) & 0.837 (3.407)   \\
\textbf{10} && 1.162 (2.439) & 0.770 (3.135)  \\
\textbf{50} && 0.896 (1.881) & 0.589 (2.398)  \\
\textbf{100} && 0.778 (1.634) & 0.498 (2.029) \\
\textbf{500} && 0.574 (1.206) & 0.294 (1.199) \\
\textbf{1,000} && 0.482 (1.106) & \textbf{0.245 (1.000)} \\
\textbf{5,000} && \textbf{0.476 (1.000)} & 0.270 (1.101) \\
\textbf{10,000} && \textbf{0.476 (1.000)} & 0.323 (1.315) \\ \midrule
\textbf{tuning} && \textbf{0.476 (1.000)} & 0.323 (1.315) \\
\bottomrule
\end{tabular}}
\end{minipage}
\vskip 0.05in
\raggedright
\fontsize{9.0pt}{9.0pt}\selectfont \textit{Note}: 
Root mean squared errors (RMSEs) estimated with 200 different bootstrapped iterations are reported ($n=300,000$).
RMSEs normalized by the best (lowest) RMSE are reported in parentheses.
$^{1/2/3/4/5/6/7}$ denote a significant difference compared to the indicated estimator (Wilcoxon rank-sum test, $p<0.05$). 
The \best{red bold} is used when the best results outperform the second bests in a statistically significant level.
The \worst{blue bold} is used when the worst results underperform the second worsts in a statistically significant level.
$\pi_b \rightarrow \pi_e$ represents the OPE situation where the estimators aim to estimate the policy value of $\pi_e$ using logged bandit data collected by $\pi_b$.
\end{table*}

\begin{table*}[t]
\begin{minipage}{1.\textwidth}
\caption{Comparison of \textit{small}-sample and \textit{large}-sample OPE performance (RMSE $\times 10^3$)} \label{tab:small_sample_ts}
\vspace{0.05in}
\large
\centering
\def\arraystretch{1.25}
\scalebox{0.6}{
\begin{tabular}{cc|cc|cc|cc}
\toprule
 && \multicolumn{2}{c|}{\textbf{\textbf{ALL}}} & \multicolumn{2}{c|}{\textbf{Men's}} & \multicolumn{2}{c}{\textbf{Women's}} \\
\cmidrule{3-8}
\textbf{OPE Estimators}  && \textbf{\textit{small}-sample} & \textbf{\textit{large}-sample} & \textbf{\textit{small}-sample} & \textbf{\textit{large}-sample} & \textbf{\textit{small}-sample} & \textbf{\textit{large}-sample} \\ 
\midrule \midrule
\textbf{IPW}$^1$ && 1.899 & 0.493$^{3}$ & 3.683$^{5}$ & 0.789 & \worst{3.156} & 0.776$^{3}$  \\
\textbf{SNIPW}$^2$ && 1.641 & 0.507$^{3}$ & 3.661$^{4/5/6}$ & 0.644$^{1/3}$ & 3.038$^{1/5}$ &  0.804$^{3}$ \\
\textbf{DM}$^3$ && 0.797$^{1/2/4/5/6}$ & \worst{1.026} & \best{3.041}$^{1/2/4/5/6/7}$ & 0.773& \best{2.665}$^{1/2/4/5/6/7}$ & \worst{0.816}$^{3}$ \\
\textbf{DR}$^4$ && 1.203$^{1/2}$ & 0.482$^{3}$ & 3.747& 0.613$^{1/3}$ & 3.055$^{1/2}$ & 0.803$^{3}$ \\
\textbf{SNDR}$^5$ && 1.159$^{1/2}$ & 0.482$^{3}$ & 3.757 & 0.659 $^{1/3}$ & 3.069 $^{1}$ & 0.791$^{3}$ \\
\textbf{Switch-DR}$^6$ && 1.203$^{1/2}$ & 0.482$^{3}$ & 3.747 & 0.613 $^{1/3}$ & 3.055 $^{5}$ & 0.803 $^{3}$ \\
\textbf{DRos}$^7$ && 0.765$^{1/2/4/5/6}$ & \best{0.316 }$^{1/2/3/4/5/6}$ & 3.727 & \best{0.459}$^{1/2/3/4/5/6}$ & 3.051$^{1/5}$  &  \best{0.561}$^{1/2/3/4/5/6}$ \\
\bottomrule
\end{tabular}}
\end{minipage}
\vskip 0.05in
\raggedright
\fontsize{9.0pt}{9.0pt}\selectfont \textit{Note}:
Root mean squared errors (RMSEs) estimated with 200 different bootstrapped iterations are reported (\textbf{Bernoulli TS $\rightarrow$ Random}).
$n=10,000$ for the \textit{small}-sample setting, while $n=300,000$ for the \textit{large}-sample setting.
$^{1/2/3/4/5/6/7}$ denote a significant difference compared to the indicated estimator (Wilcoxon rank-sum test, $p<0.05$). 
The \best{red bold} is used when the best results outperform the second bests in a statistically significant level.
The \worst{blue bold} is used when the worst results underperform the second worsts in a statistically significant level.
\end{table*}

\subsection{Results}
The results of the benchmark experiments with $n=300,000$ are given in Table~\ref{tab:se_ts}.
We describe \textbf{Bernoulli TS} $\rightarrow$ \textbf{Random} to represent the OPE situation where we use Bernoulli TS as $\pi_b$ and Random as $\pi_e$. 
Please see Appendix~\ref{app:experimental_settings} for additional results.

\paragraph{Performance comparisons.}
Table~\ref{tab:se_ts} shows that DRos (with automatic hyperparameter tuning) performs best for the three campaigns, achieving about 30-60\% more accurate OPE than the second-best estimators.
We then evaluate several values for the hyperparameter $\lambda$ of DRos.
Table~\ref{tab:hyperparam} shows the OPE performance (RMSE) of DRos with different values of $\lambda$ ($\in \{1,5,10,50,\ldots,10000\}$).
This table also includes the OPE performance of DRos with automatic hyperparameter tuning of Su et al.~\citep{su2020doubly}.
First, we observe that the choice of $\lambda$ greatly affects the performance of DRos.
Specifically, for \textbf{Random $\rightarrow$ Bernoulli TS}, a larger value of $\lambda$ leads to a better OPE performance.
In contrast, for \textbf{Bernoulli TS $\rightarrow$ Random}, $\lambda=1,000$ is the best setting.
Second, we observe that the automatic hyperparameter tuning procedure prefers a large value of $\lambda$.
This means that the tuning procedure puts emphasis on the bias of the estimator, as a large value of $\lambda$ leads to a low bias, but a high variance estimator.
This strategy succeeds for \textbf{Random $\rightarrow$ Bernoulli TS}.
However, we observe that there is room for improvement for \textbf{Bernoulli TS $\rightarrow$ Random} in terms of automatic hyperparameter tuning.
This suggests opportunities for further investigations on the quality of automatic hyperparameter tuning for achieving more accurate OPE in practice.

\paragraph{OPE performance with different sample sizes.}
Next, we compare the \textit{small}-sample setting ($n=10,000$) and \textit{large}-sample setting ($n=300,000$) to evaluate how the OPE performance changes with the sample size.
We observe in Table~\ref{tab:small_sample_ts} that the estimators' performance can change significantly depending on the size of the logged bandit data.
In particular, for the Men's and Women's campaigns, the most accurate estimator changes with the sample size.
The table shows that DM outperforms the other estimators in the \textit{small}-sample setting, while DRos is the best for the \textit{large}-sample setting.
These observations suggest that practitioners have to choose an appropriate OPE estimator carefully for their specific application.
It is thus necessary to develop a reliable method to choose and tune OPE estimators in a data-driven manner.
Specifically, in real applications, we have to tune the estimators' hyperparameters or identify an accurate estimator without the ground-truth or on-policy policy value of the evaluation policy.

\section{Conclusion,  Limitations, and Future Work} \label{sec:conclusion}
To enable a realistic and reproducible evaluation of OPE, we presented Open Bandit Dataset, a set of logged bandit datasets collected on a fashion e-commerce platform.
The dataset comes with Open Bandit Pipeline, Python software that makes it easy to evaluate and compare different OPE estimators.
We aim to facilitate understanding of the empirical properties of OPE estimators and address experimental inconsistencies in the literature.
We also perform extensive experiments on a variety of OPE estimators and analyze the effects of hyperparameter choice and sample size in a real-world setting.
Our experiments highlight that an appropriate estimator can change depending on a problem setting such as the sample size.
The results also suggest that there is room to improve the OPE performance in terms of automatic hyperparameter tuning and estimator selection.
These observations call for a new estimator selection method and a hyperparameter tuning procedure to be developed.

A limitation is that we assume that the reward of an item at a position does not depend on other simultaneously presented items.
This assumption might not hold, as an item's attractiveness can have a significant effect on the expected reward of another item in the same recommendation list~\citep{li2018offline}. 
To address more realistic situations, we have implemented some OPE estimators for the slate action setting~\cite{mcinerney2020counterfactual,swaminathan2017off} in Open Bandit Pipeline.
Comparing the standard OPE estimators and those for the slate action setting on our dataset is a valuable and interesting research direction.

Open Bandit Dataset is currently the only public dataset allowing OPE experiments.
Therefore, it might lead to an overfitting issue.
Moreover, Open Bandit Dataset includes only two policies.
Here, we emphasize that there has never been any public real-world data that allow realistic and reproducible OPE research before. 
Our open-source project is an initial step towards this goal. 
Having many policies would be even more valuable, but releasing data with two different data collection policies is distinguishable enough from the prior work.
We believe that our work will inspire other researchers and companies to create follow-up benchmark datasets to advance OPE research further.

\begin{ack}
We thank Haruka Kiyohara, Ryo Kuroiwa, Richard Liu, Kazuki Mogi, Masahiro Nomura, Kyohei Okumura, Ayumi Sudo, and Koichi Takayama for their thoughtful comments on the manuscript and help in developing the software. We would also like to thank anonymous reviewers for their helpful feedback. 
\end{ack}

\bibliographystyle{plain}
\bibliography{main}

\clearpage
\tableofcontents

\clearpage
\appendix
\section{Examples of Bandit Algorithms} \label{app:examples}
Our setup allows for many popular multi-armed bandit algorithms and off-policy learning methods, as the following examples illustrate.

\begin{example}[Random A/B testing]\label{ex:A/B}
	We always choose each action uniformly at random, i.e., 
	$\pi_{\mathrm{Uniform}}(a \mid x) =|\calA|^{-1}$ 
	always holds for any given $a \in \calA$ and $x \in \calX$.
\end{example}
\vspace{0.2in}
\begin{example}[Bernoulli Thompson Sampling]\label{ex:ts}
	We sample the potential reward $\tilde{r}(a)$ from the beta distribution $Beta (S_{ta} + \alpha, F_{ta} + \beta) $ for each action in $\calA$, where $S_{ta} := \sum_{t'=1}^{t-1} r_{t'}, F_{ta} := (t-1) - S_{ta}$. $(\alpha, \beta)$ are the parameters of the prior Beta distribution. We then choose the action with the highest sampled potential reward, $a: \in \argmax_{a^{\prime} \in \calA} \tilde{r}(a^{\prime})$ (ties are broken arbitrarily). 
	As a result, this algorithm chooses actions with the following probabilities:
    $$ 
    \pi_{\mathrm{BernoulliTS}}(a \mid x) = \Pr\{a \in \argmax_{a^{\prime} \in \calA} \tilde{r}(a^{\prime})\}
    $$
    for any given $a \in \calA$ and $x \in \calX$. When implementing the data collection experiment on the ZOZOTOWN platform, we modified TS to adjust to our top-3 recommendation setting shown in Figure~\ref{fig:displayed_fashion_item_sample}. The modified TS selects three actions with the three highest sampled rewards which create a nonrepetitive set of item recommendations for each coming user.
\end{example}
\vspace{0.2in}
\begin{example}[IPW Learner]\label{ex:ts}
	When $\calD$ is given, we can train a deterministic policy $\pi_{\mathrm{det}}: \calX \rightarrow \calA$ by maximizing the IPW estimator as follows:
    \begin{align*}
        \pi_{\mathrm{det}}(x) 
        & \in  \argmax_{\pi \in \Pi} \hat{V}_{\mathrm{IPW}}(\pi ; \mathcal{D}) \\
        & =\argmax_{\pi \in \Pi} \mathbb{E}_{\mathcal{D}}\left[\frac{\mathbb{I}\left\{\pi\left(x_{t}\right)=a_{t}\right\}}{\pi_{b}\left(a_{t} \mid x_{t}\right)} r_{t}\right] \\
        & =\argmin_{\pi \in \Pi} \mathbb{E}_{\mathcal{D}}\left[\frac{r_i}{\pi_{b}\left(a_{t} \mid x_{t}\right)} \mathbb{I}\left\{\pi\left(x_{t}\right) \neq a_{t}\right\}\right] 
    \end{align*}
    , which is equivalent to the cost-sensitive classification problem that can be solved with an arbitrary machine learning classifier.
\end{example}

\clearpage
\section{OPE Estimators and Related Techniques} \label{app:estimators}
Here we define some advanced OPE estimators compared in Section~\ref{sec:benchmark_experiment}.

\paragraph{Self-Normalized Estimators.}
Self-Normalized Inverse Probability Weighting (SNIPW) is an approach to address the variance issue with the original IPW.
It estimates the policy value by dividing the sum of weighted rewards by the sum of importance weights as follows.
\begin{align*}
    \snipw :=\frac{\mE_{n} [ w(x_i,a_i) r_i ]}{\mE_{n} [ w(x_i,a_i) ]}.
\end{align*}
SNIPW is more stable than IPW, because the policy value estimated by SNIPW is bounded in the support of rewards and its conditional variance given action and context is bounded by the conditional variance of the rewards~\citep{kallus2019intrinsically}.
IPW does not have these properties.
We can define Self-Normalized Doubly Robust (SNDR) in a similar manner as follows.
\begin{align*}
    \sndr := \mE_{n} \left[\hat{q}(x_i, \pi_e) + \frac{w(x_i,a_i) }{\mE_{n} [ w(x_i,a_i) ]} (r_i-\hat{q}(x_i, a_i) ) \right].
\end{align*}

\paragraph{Switch Estimator.}
The original DR estimator can still be subject to the variance issue, particularly when the importance weights are large due to weak overlap.
Switch-DR aims to alleviate the variance issue by using DM where the importance weights are large as:
\begin{align*}
    \switchdr 
    := \mE_{n} \left[ \hat{q}(x_i, \pi_e) + w(x_i,a_i) (r_i-\hat{q}(x_i, a_i) ) \mathbb{I}\{ w(x_i,a_i) \le \tau \} \right],
\end{align*}
where $\mathbb{I} \{\cdot\}$ is the indicator function and $\tau \ge 0$ is a hyperparameter.
Switch-DR interpolates between DM and DR. 
When $\tau=0$, it is identical to DM, while $ \tau \to \infty $ yields DR.

\paragraph{Doubly Robust with Optimistic Shrinkage (DRos).}
Su et al.~\citep{su2020doubly} propose DRos based on a new weight function $\hat{w}: \mathcal{X} \times \mathcal{A} \rightarrow \mathbb{R}_{+}$, which directly minimizes sharp bounds on the MSE of the resulting estimator. 
DRos is defined as 
\begin{align*}
    \drs := \mE_{n} [ \hat{q} (x_i, \pi_e) + \hat{w} (x_i, a_i; \lambda)  (r_i-\hat{q}(x_i, a_i) ) ],
\end{align*}
where $\lambda \ge 0$ is a pre-defined hyperparameter and the new weight is
\begin{align*}
    \hat{w} (x, a; \lambda) := \frac{\lambda}{w^{2}(x, a)+\lambda} w(x, a).
\end{align*}
When $\lambda = 0$, $\hat{w} (x, a; \lambda) = 0$ leads to DM. 
On the other hand, as $\lambda \rightarrow \infty$, $\hat{w} (x, a; \lambda) = w(x,a)$ leading to DR.

\paragraph{Cross-Fitting Procedure.}
We sometimes use \textit{cross-fitting} to avoid the overfitting issue in obtaining $\hat{q}$ from logged bandit data~\citep{narita2020safe}.
We describe the cross-fitting procedure below.
\begin{enumerate}
    \item Take a $K$-fold random partition $\left(\calD_{k}\right)_{k=1}^{K}$ of size $n$ of $\calD$ such that the size of each fold is $n_k = n / K$. We also define $\calD_{k}^{c}:=\calD \backslash \calD_{k}$ for $k=1,2,\ldots K$.
    \item For each $k=1,2,\ldots K$, obtain reward estimators $\{\hat{q}_k\}_{k=1}^K$ with $\calD_{k}^{c}$.
    \item Given reward estimators $\{\hat{q}_k\}_{k=1}^K$, estimate $V(\pi_e)$ by
    $ K^{-1} \sum_{k=1}^K \hat{V} (\pi_e; \calD_{k}, \hat{q}_k) $ where $\hat{V}$ is a model-dependent estimator.
\end{enumerate}
$K (\ge 2)$ is the number of folds and we use $K=2$ in the benchmark experiments.

\paragraph{Automatic Hyperparameter Tuning.}
As we have summarized, many estimators in the OPE community depend on hyperparameters such as $\lambda$ and $\tau$.
Recently, an automatic hyperparameter tuning procedure is proposed by Su et al.~\citep{su2020doubly}, which tunes the hyperparameter values as follows.
\begin{align*}
    \hat{\theta} \in \argmin_{\theta \in \Theta} \, \overline{\mathrm{Bias}}(\theta; \calD)^2 + \hat{\mathbb{V}} (\theta; \calD),
    \label{eq:hyperparameter_iuning}
\end{align*}
where $\overline{\mathrm{Bias}}(\theta; \calD)$ is the bias upper bound estimable with $\calD$, and $\hat{\mathbb{V}} (\theta; \calD)$ is the sample variance. 
Su et al.~\citep{su2020doubly} describe some variants of the bias upper bound including the following direct bias estimation:
\begin{align*}
    \overline{\mathrm{Bias}}(\theta; \calD) 
    & := \left|\mE_{n} [ \left(\hat{w}\left(x_{i}, a_{i}; \theta \right)-w\left(x_{i}, a_{i}\right)\right) \left(r_{i}-\hat{q}\left(x_{i}, a_{i}\right)\right) ] \right| \\
    & +\sqrt{\frac{2 \mathbb{E}\left[w (x, a)^{2}\right] \log (2 / \delta)}{n}}+\frac{2 w_{\max} \log (2 / \delta)}{3 n}
\end{align*}
where $\delta \in (0, 1]$ defines a confidence in deriving a high probability upper bound, and $w_{\max } := \max_{x,a} w (x,a)$ is the maximum importance weight.
$\hat{w} (x, a; \theta )$ is the importance weight modified by a hyperparameter.
For example, DRos use $\hat{w} (x, a; \lambda) = \frac{\lambda}{w^{2}(x, a)+\lambda} w(x, a)$, while Switch-DR is based on $\hat{w} (x, a; \tau ) = w(x, a) \mathbb{I}\{ w(x, a) \le \tau \}$.
In Section~\ref{sec:benchmark_experiment}, we use this direct bias estimation with $\delta=0.05$ to conduct automatic hyperparameter tuning.

\clearpage
\begin{algorithm*}[t]
\caption{Experimental protocol for evaluating OPE estimators} 
\label{algo:evaluation_of_ope}
\begin{algorithmic}[1]
\Require policy $\pi_e$; two different logged bandit datasets $\calD^{(e)} = \{ (x^{(e)}_i, a^{(e)}_i, r^{(e)}_i) \}_{i=1}^{n^{(e)}}$ and $\calD^{(b)} = \{ (x^{(b)}_i, a^{(b)}_i, r^{(b)}_i) \}_{i=1}^{n^{(b)}}$ where $\calD^{(e)}$ is collected by $\pi_e$ and $\calD^{(b)}$ is collected by a different one $\pi_b$; an off-policy estimator to be evaluated $\hat{V}$; a number of bootstrap iterations $T$; sample size $n$
\Ensure some statistics of $\textit{SE} (\hat{V})$ (such as the root mean squared error)
\State $\onpolicy = (1/n^{(e)}) \sum_{i=1}^{n^{(e)}} r^{(e)}_i$  (on-policy estimation of $V(\pi_e)$)
\For {$t=1, \ldots, T$}
    \State $\calD^{(b,*)}_{t} = \operatorname{Bootstrap}(\calD^{(b)}, n)$ (sample data of size $n$ from $\calD^{(b)}$ with \textit{replacement})
    \State $\textit{SE} (\hat{V}; \calD^{(b)}) := \left( \hat{V}(\pi_e; \calD^{(b)}) -  \onpolicy \right)^2,$ (calculate the squared error in OPE)
    \State $\mathcal{S} \leftarrow \mathcal{S} \cup \{ \textit{SE} (\hat{V}; \calD^{(b,*)}_{\mathrm{ev}}) \} $ 
\EndFor
\end{algorithmic}
\end{algorithm*}

\begin{table*}[ht]
\begin{minipage}{1.0\textwidth}
\caption{Elapsed Time (minutes)} \label{tab:elapsed_time}
\vskip 0.05in
\large
\centering
\def\arraystretch{1.25}
\scalebox{0.75}{
\begin{tabular}{c|cccccc}
\toprule
 \textbf{Sample Size ($n$)} && \textbf{ALL} && \textbf{Men's} && \textbf{Women's} \\
\midrule \midrule
 \textbf{10,000} && 22.38  && 10.22 && 12.96 \\
 \textbf{30,000} && 48.73  && 19.23 && 24.18 \\
 \textbf{50,000} && 107.84  && 29.00 && 39.25 \\
 \textbf{100,000} && 208.91 && 57.58 && 89.70 \\
 \textbf{300,000} && 750.34  && 220.53 && 332.56 \\
\bottomrule
\end{tabular}}
\vskip 0.05in
\raggedright
\fontsize{9.0pt}{9.0pt}\selectfont \textit{Note}: 
This table presents the time (in minutes) needed to run the evaluation of OPE experiment using Algorithm~\ref{algo:evaluation_of_ope} with $T=200$ and different sample size $n \in \{10000,30000,50000,100000,300000\}$.
\end{minipage}
\\
\vspace{0.2in}
\\
\begin{minipage}{1.0\textwidth}
\caption{Estimation performance of reward estimator $\hat{q}$ with different machine learning methods} \label{tab:q_hat_acc}
\vskip 0.05in
\large
\centering
\def\arraystretch{1.3}
\scalebox{0.65}{
\begin{tabular}{c|cccccccc}
\toprule
 & \multicolumn{3}{c}{\textbf{Random $\rightarrow$ Bernoulli TS}} && \multicolumn{3}{c}{\textbf{Bernoulli TS $\rightarrow$ Random}} \\
 \cmidrule{2-4} \cmidrule{6-8}
 \textbf{Models ($\hat{q}$)} & \textbf{ALL} & \textbf{Men's} & \textbf{Women's} && \textbf{ALL} & \textbf{Men's} & \textbf{Women's} \\
\midrule \midrule
 \textbf{GB} & \std{0.2954}{0.1000} & \std{0.1871}{0.0756} & \std{0.2598}{0.0555} && \std{0.0854}{0.0751} & \std{0.1028}{0.1337} & \std{0.0910}{0.0422} \\
 \textbf{LR} & \std{0.1091}{0.0313} & \std{0.0745}{0.0194} & \std{0.0821}{0.0250} && \std{0.0338}{0.0145} & \std{0.0641}{0.0202} & \std{0.0232}{0.0110} \\
\bottomrule
\end{tabular}
}
\vskip 0.05in
\raggedright
\fontsize{8.5pt}{8.5pt}\selectfont \textit{Note}: 
This table presents the relative cross-entropy (RCE) of the reward estimator for each campaign.
The averaged results and their unbiased standard deviations estimated using 200 different bootstrapped samples are reported.
\textbf{GB} and \textbf{LR} stand for Gradient Boosting and Logistic Regression, respectively.
$\pi_b \rightarrow \pi_e$ represents the OPE situation where the estimators aim to estimate the policy value of $\pi_e$ using logged bandit data collected by $\pi_b$, meaning that $\hat{q}$ is trained on data collected by $\pi_b$.
\end{minipage}
\end{table*}

\section{Additional Experimental Setups and Results} \label{app:experimental_settings}
In this section, we describe some additional experimental setups and results.

\subsection{Computational Resource}
All experiments were conducted on MacBook Pro (2.4 GHz Intel Core i9, 64 GB). 
Table~\ref{tab:elapsed_time} summarizes the time needed to run the evaluation of OPE experiments with $T=200$ and varying sample size $n \in \{10000,30000,50000,100000,300000\}$.


\subsection{Estimation Performance of Reward Estimator ($\hat{q}$)}
We evaluate the performance of the reward estimators by using relative cross entropy (RCE).
RCE is defined as the improvement of an estimation performance relative to the naive estimation, which uses the mean CTR for every prediction.
We calculate this metric using a size $n$ of validation samples $\{ (x_i, a_i, r_i) \}_{i=1}^n$ as:
\begin{align*}
    \textit{RCE}\ (\hat{q}) := 1 - \frac{\sum_{i=1}^n r_i \log( \hat{q} (x_i,a_i) ) + (1-r_i) \log(1-\hat{q} (x_i,a_i))}{\sum_{i=1}^n r_i \log(\hat{q}_{\mathrm{naive}} ) + (1-r_i) \log(1-\hat{q}_{\mathrm{naive}})}
\end{align*}
where $\hat{q}_{\mathrm{naive}} := n^{-1} \sum_{i=1}^n r_i $ is the naive estimation using the mean CTR for every estimation.
A larger value of RCE means a better performance of a predictor.
We use RCE as an evaluation metric to evaluate $\hat{q}$, as in OPE, the estimation accuracy of $\hat{q}$ (not Recall or Precision) matters.

Table~\ref{tab:q_hat_acc} reports the estimation accuracy of logistic regression and gradient boosting as a reward estimator $\hat{q}$.
Specifically, we use `sklearn.linear\_model.LogisticRegression(C=100, random\_state=12345)' as logistic regression and `sklearn.ensemble.HistGradientBoostingClassifier(learning\_rate=0.01, max\_iter=100, max\_depth=5, min\_samples\_leaf=10, random\_state=12345)' as gradient boosting to obtain $\hat{q}$.
Note that we use action-related feature vectors to represent action variables to train reward estimators.
The action-related feature vectors used in the benchmark experiments are available at \url{https://github.com/st-tech/zr-obp/blob/master/obd/bts/all/item\_context.csv}.
Note also that, in the evaluation of OPE experiments, we use gradient boosting (with cross-fitting) to obtain $\hat{q}$.
Table~\ref{tab:q_hat_acc} contains the estimation accuracy of logistic regression as a baseline.
The table shows that gradient boosting is more accurate in estimating $q$ than logistic regression.

\subsection{Additional Experimental Results}
Table~\ref{tab:se_ur} shows the benchmark results (RMSE) for the three campaigns when the OPE situation is \textbf{Random $\rightarrow$ Bernoulli TS}.
Table~\ref{tab:small_sample_ur} compares \textbf{\textit{small}}-sample and \textbf{\textit{large}}-sample OPE performance on the three campaigns when the OPE situation is \textbf{Random $\rightarrow$ Bernoulli TS}.
Table~\ref{tab:different_q_ts} compares the OPE performance of the model-dependent estimators (i.e., DM, DR, SNDR, Switch-DR, and DRos) with different reward estimators. 
The results show that the choice of machine learning method to construct $\hat{q}$ greatly affects the OPE performance.
Specifically, gradient boosting leads to a better OPE than logistic regression as its reward estimation is relatively accurate (see Table~\ref{tab:q_hat_acc}).
We also observe the similar trend for Men's and Women's campaigns. \\

\begin{table}[h]
\begin{minipage}{1.0\textwidth}
\caption{RMSE ($\times 10^3$) of OPE estimators (\textbf{Random $\rightarrow$ Bernoulli TS})} \label{tab:se_ur}
\vskip 0.05in
\large
\centering
\def\arraystretch{1.25}
\scalebox{0.65}{
\begin{tabular}{c|cccccc}
\toprule
 \textbf{OPE Estimators ($\hat{V}$)} && \textbf{ALL} && \textbf{Men's} && \textbf{Women's} \\
\midrule \midrule
 \textbf{IPW}$^1$ && 0.474 (1.012)$^{3}$ && 1.298 (1.145)$^{3}$ && 1.645 (2.817) \\
 \textbf{SNIPW}$^2$ && 0.468 (1.000)$^{3}$ && 1.133 (1.000)$^{1/3}$ && 1.585 (2.713) \\
 \textbf{DM}$^3$ && \worst{1.499 (3.201)} && \worst{1.758 (1.551)} && \best{0.584 (1.000)}$^{1/2/4/5/6/7}$ \\
 \textbf{DR}$^4$ && 0.501 (1.070)$^{3}$ && 1.169 (1.032)$^{1/3}$ && 1.539 (2.636)$^{1}$ \\
 \textbf{SNDR}$^5$ && 0.500 (1.067)$^{3}$  && 1.151 (1.032)$^{1/3}$ && 1.524 (2.609)$^{1}$ \\
 \textbf{Switch-DR}$^6$ && 0.501 (1.070)$^{3}$  && 1.169 (1.032)$^{1/3}$ && 1.539 (2.036)$^{1}$ \\
 \textbf{DRos}$^7$ && 0.508 (1.085)$^{3}$  && 1.171 (1.033)$^{1/3}$ && 1.524 (2.609)$^{1}$ \\
\bottomrule
\end{tabular}}
\end{minipage}
\\
\vspace{0.2in}
\\
\begin{minipage}{1.\textwidth}
\caption{Comparison of \textit{small}-sample and \textit{large}-sample OPE performance (RMSE $\times 10^3$)} \label{tab:small_sample_ur}
\vspace{0.05in}
\large
\centering
\def\arraystretch{1.25}
\scalebox{0.6}{
\begin{tabular}{cc|cc|cc|cc}
\toprule
 && \multicolumn{2}{c|}{\textbf{\textbf{ALL}}} & \multicolumn{2}{c|}{\textbf{Men's}} & \multicolumn{2}{c}{\textbf{Women's}} \\
\cmidrule{3-8}
\textbf{OPE Estimators}  && \textbf{\textit{small}-sample} & \textbf{\textit{large}-sample} & \textbf{\textit{small}-sample} & \textbf{\textit{large}-sample} & \textbf{\textit{small}-sample} & \textbf{\textit{large}-sample} \\ 
\midrule \midrule
\textbf{IPW}$^1$ && 1.104 & 0.474$^{3}$ & 3.156 & 1.298$^{3}$ & \worst{6.789} & 1.645   \\
\textbf{SNIPW}$^2$ && 1.100$^{3/5}$ & 0.468$^{3}$ & 3.038 & \textbf{1.133}$^{1/3}$ & 5.685$^{1}$ & 1.585  \\
\textbf{DM}$^3$ && 1.128 & \worst{1.499} & \best{2.665}$^{1/2/4/5/6/7}$ & \worst{1.758} & \best{2.986}$^{1/2/4/5/6/7}$  & \best{0.584}$^{1/2/4/5/6/7}$  \\
\textbf{DR}$^4$ && 1.133 & 0.501$^{3}$ & 3.055 & 1.169$^{1/3}$ & 5.724$^{1}$  & 1.539$^{1}$  \\
\textbf{SNDR}$^5$ && 1.148 & 0.500$^{3}$ & 3.069 & 1.151$^{1/3}$ & 5.476$^{1}$ & 1.524$^{1}$ \\
\textbf{Switch-DR}$^6$ && 1.133 & 0.501$^{3}$ & 3.055 & 1.169$^{1/3}$ & 5.724$^{1}$ & 1.539$^{1}$ \\
\textbf{DRos}$^7$ && 1.125 & 0.500$^{3}$ & 3.051 & 1.171$^{1/3}$ & 5.694$^{1}$ & 1.524$^{1}$ \\
\bottomrule
\end{tabular}}
\end{minipage}
\vskip 0.05in
\raggedright
\fontsize{9.0pt}{9.0pt}\selectfont \textit{Note}:
Root mean squared errors (RMSEs) estimated with 200 different bootstrapped iterations are reported (\textbf{Random $\rightarrow$ Bernoulli TS}).
RMSEs normalized by the best (lowest) RMSE are reported in parentheses.
$n$ is the sample size of $\calD^{(b,*)}$, which is used in OPE.
$n=10,000$ for the \textit{small}-sample setting, while $n=300,000$ for the \textit{large}-sample setting.
$^{1/2/3/4/5/6/7}$ denote a significant difference compared to the indicated estimator (Wilcoxon rank-sum test, $p<0.05$). 
The \best{red bold} is used when the best results outperform the second bests in a statistically significant level.
The \worst{blue bold} is used when the worst results underperform the second worsts in a statistically significant level.
\\
\vspace{0.2in}
\begin{minipage}{1.\textwidth}
\caption{Comparison of OPE performance (RMSE $\times 10^3$) with different reward estimators} \label{tab:different_q_ts}
\vspace{0.05in}
\large
\centering
\def\arraystretch{1.25}
\scalebox{0.65}{
\begin{tabular}{cc|cc|c}
\toprule
\textbf{OPE Estimators}  && \textbf{LR} & \textbf{GB} & \textbf{Improvement of GB over LR} \\
\midrule \midrule
\textbf{DM} && 1.597 & 1.026 & 64.2\% \\
\textbf{DR} && 2.250 & 0.482 & 78.5\% \\
\textbf{SNDR} && 2.255 & 0.482 & 78.7\% \\
\textbf{Switch-DR} && 2.250 & 0.482 & 78.5\% \\
\textbf{DRos} && 2.065 & 0.316 & 84.7\% \\
\bottomrule
\end{tabular}
}
\end{minipage}
\vskip 0.05in
\raggedright
\fontsize{9.0pt}{9.0pt}\selectfont \textit{Note}:
Root mean squared errors (RMSEs) estimated with 200 different bootstrapped iterations are reported ($n=300,000$, \textbf{ALL} campaign, \textbf{Bernoulli TS $\rightarrow$ Random}).
\textbf{GB} and \textbf{LR} stand for Gradient Boosting and Logistic Regression, respectively.
\end{table}

\clearpage

\vspace{5mm}
\begin{figure*}[t]
    \centering
    \includegraphics[clip, width=13cm]{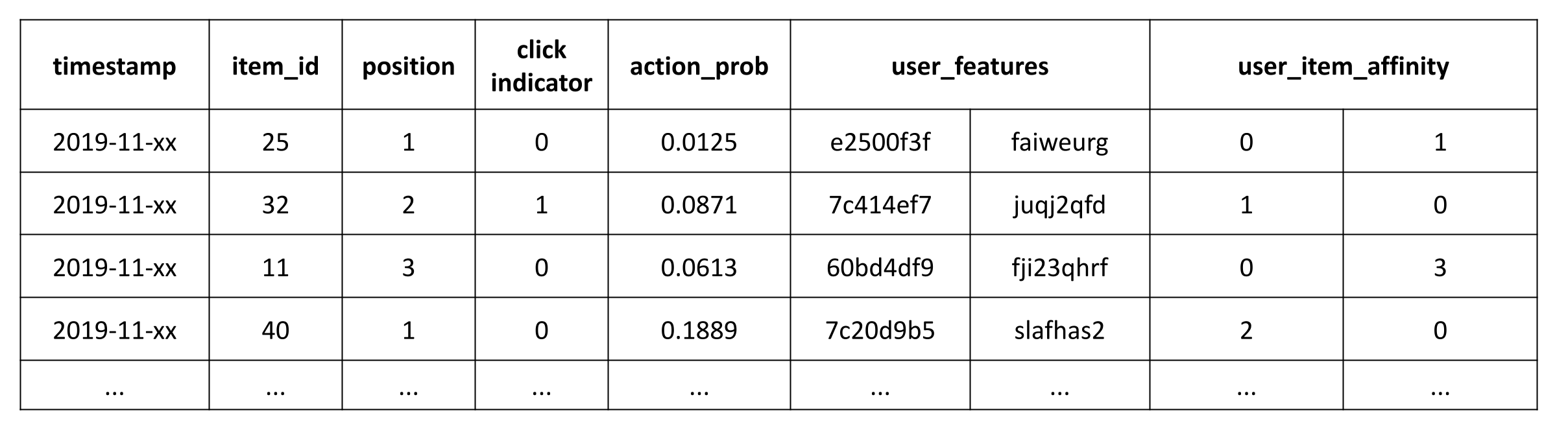}
    \caption{Schema of Open Bandit Dataset.}
    \label{fig:schema}
\end{figure*}

\section{Open Bandit Dataset} \label{app:obd}

We publish \textit{Open Bandit Dataset} at: \url{https://research.zozo.com/data.html}. 
It is our priority to protect the privacy of third parties.
We bear all responsibility in case of violation of rights, etc., and confirmation of the data license.

\paragraph{Overview}
The dataset currently consists of a total of about 26M rows, each representing a user impression with some feature values, selected items as actions, true action choice probabilities by the data collection policies, and click indicators as reward variables.
Specifically, Figure~\ref{fig:schema} describes the schema of Open Bandit Dataset, where
\begin{itemize}
    \item timestamp: timestamp of impressions.
    \item item\_id: index of items as arms (index ranges from 0-79 in "ALL" campaign, 0-33 for "Men" campaign, and 0-45 "Women" campaign).
    \item position: the position of an item being recommended (1, 2, or 3 correspond to the left, center, and right positions of the ZOZOTOWN recommendation interface in Figure~\ref{fig:displayed_fashion_item_sample}, respectively).
    \item click\_indicator: a reward variable that indicates if an item was clicked (1) or not (0).
    \item action\_prob: the probability of an item being recommended at the given position by a data collection policy.
    \item user features (categorical): user-related feature values such as age and gender. User features are anonymized using a hash function.
    \item user-item affinity (numerical): user-item affinity scores induced by the number of past clicks observed between each user-item pair.
\end{itemize}

\paragraph{Dataset Documentation}
We include the dataset documentation and intended uses in the dataset website.

\paragraph{Terms of use and License.}
Open Bandit Dataset is published under CC BY 4.0. license, which means everyone can use this dataset for non-commercial research purposes.

\paragraph{Dataset maintenance.}
We will maintain the dataset for a long time at the same website and check the data accessibility in a regular basis. We will also maintain the small size example data at the Open Bandit Pipeline repository: \url{https://github.com/st-tech/zr-obp/tree/master/obd}. This will help potential users grasp how the dataset works with Open Bandit Pipeline. We also maintain the data loader class implemented in Open Bandit Pipeline.\footnote{https://github.com/st-tech/zr-obp/blob/master/obp/dataset/real.py} When a function of the package is updated, we will let the users know the changes via the project mailing list.

\paragraph{Benchmark and code.}
\url{https://github.com/st-tech/zr-obp/tree/master/benchmark/ope} provides scripts and instructions to reproduce the benchmark results using Open Bandit Dataset. We utilize tools such as \textit{Hydra}\footnote{https://hydra.cc/} and \textit{Poetry}\footnote{https://python-poetry.org/} to ensure the reproducibility of the benchmark results.

\paragraph{Dataset statistics.}
Figure~\ref{fig:action_dist_vs} shows the empirical action distributions in Open Bandit Dataset.
We see that the action distribution is uniform when the dataset is collected under the Random policy (see also Figure~\ref{fig:action_dist_ur}).
In contrast, we can see that some actions are much more frequently observed than others when the dataset is collected under the Bernoulli TS policy, producing the bias in OPE.

\subsection*{Potential Negative Societal Impacts and General Ethical Conduct}

Our open data and pipeline contribute to fair and transparent machine learning research, especially  bandit algorithms and off-policy evaluation. 
By setting up a common ground for credibly evaluating the performance of bandit and off-policy evaluation methods, our work is expected to foster their real-world applications. 
A limitation is that it is difficult to generalize the experimental results and conclusions based on our data to other important domains, such as education, healthcare, and the social sciences. 
To enable generalizable comparison and evaluation of bandit algorithms and off-policy evaluation, it is desired to construct public benchmark datasets from a broader range of domains.

As we have touched on in Section~\ref{sec:dataset}, we hashed the feature vectors related to the users included in the dataset. 
Therefore, our dataset does not contain any personally identifiable information or sensitive personally identifiable information.

\begin{table}[t]
\centering
\begin{minipage}{1.0\textwidth}
    \begin{center}
        \begin{tabular}{c}
            \begin{minipage}{\hsize}
                \begin{center}
                    \includegraphics[clip, width=6.0cm]{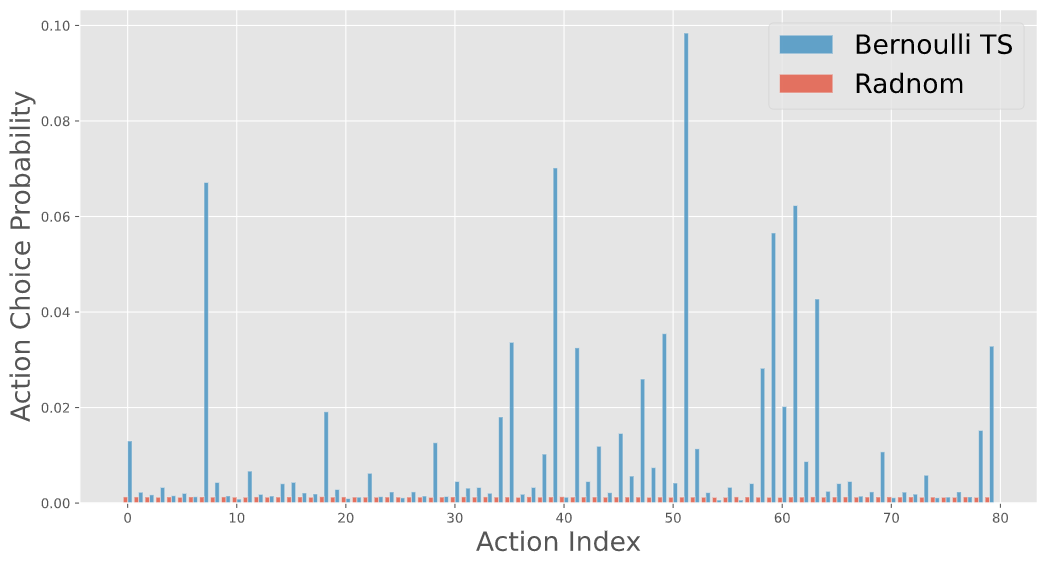}\\
                    ALL
                \end{center}
            \end{minipage}\\ \vspace{0.05in} \\
            
            \begin{minipage}{0.45\hsize}
                \begin{center}
                    \includegraphics[clip, width=5.5cm]{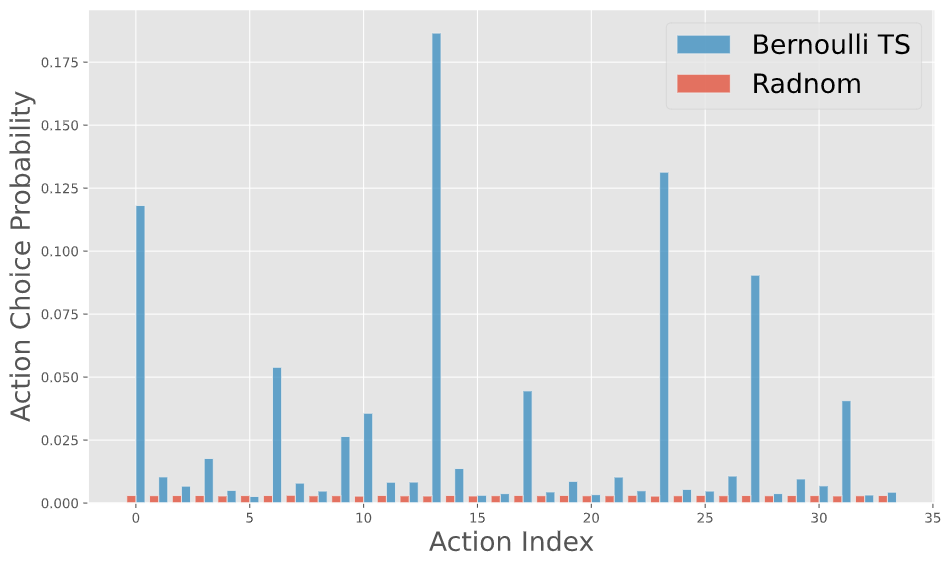}\\
                    Men's
                \end{center}
            \end{minipage}
            
            \begin{minipage}{0.45\hsize}
                \begin{center}
                    \includegraphics[clip, width=5.5cm]{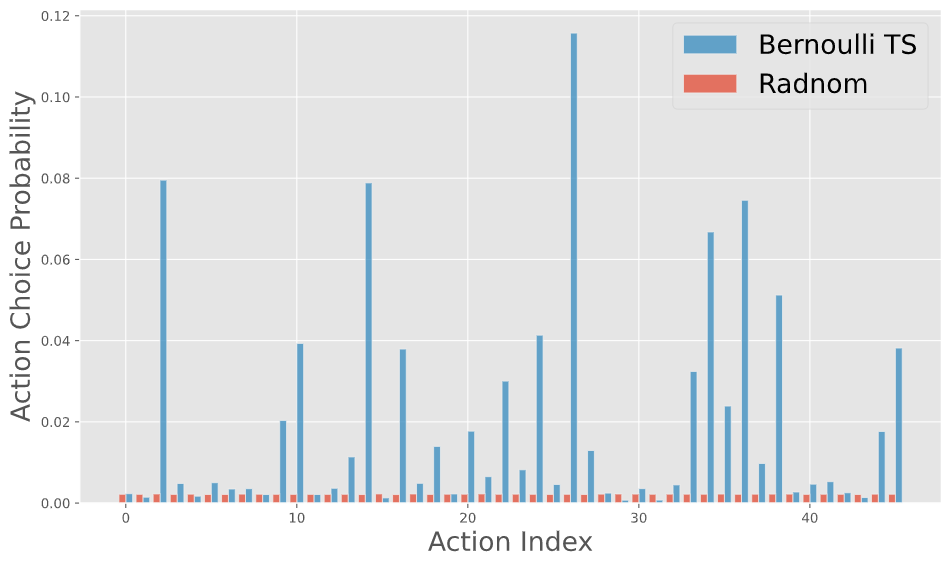}\\
                    Women's
                \end{center}
            \end{minipage}
        \end{tabular}
    \end{center}
\caption{Empirical Action Distribution (\textbf{Bernoulli TS vs Random})}
\label{fig:action_dist_vs}
\end{minipage}
\vspace{0.2in}
\begin{minipage}{1.0\textwidth}
    \begin{center}
        \begin{tabular}{c}
            \begin{minipage}{0.33\hsize}
                \begin{center}
                    \includegraphics[clip, width=4.5cm]{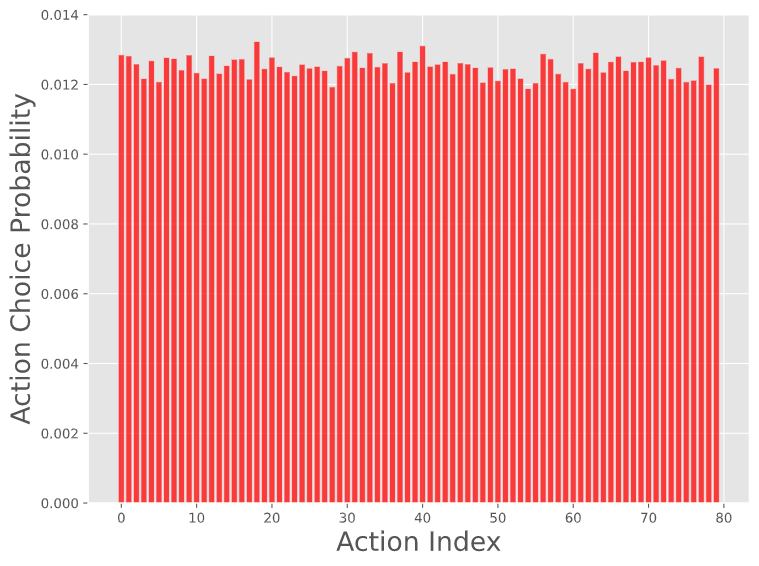}\\
                    ALL
                \end{center}
            \end{minipage}
            
            \begin{minipage}{0.33\hsize}
                \begin{center}
                    \includegraphics[clip, width=4.5cm]{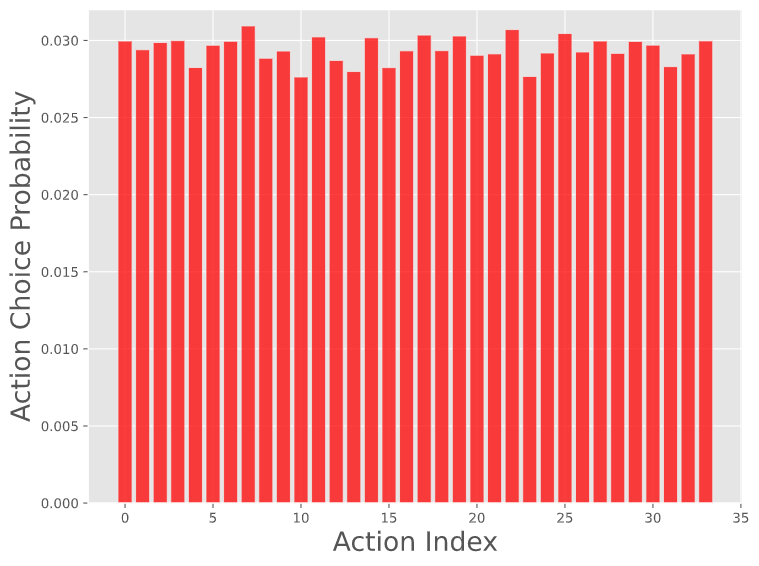}\\
                    Men's
                \end{center}
            \end{minipage}
            
            \begin{minipage}{0.33\hsize}
                \begin{center}
                    \includegraphics[clip, width=4.5cm]{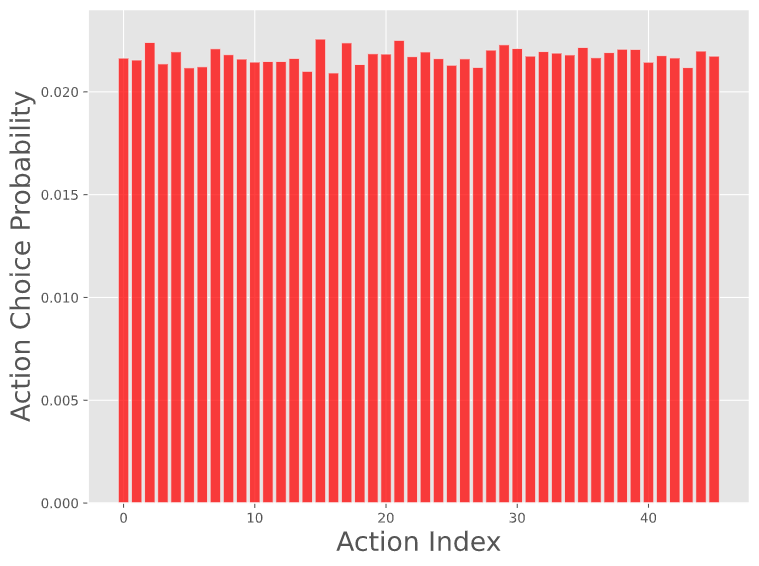}\\
                    Women's
                \end{center}
            \end{minipage}
        \end{tabular}
    \end{center}
    \caption{Empirical Action Distribution (\textbf{Random only})}
\label{fig:action_dist_ur}
\end{minipage}
\end{table}

\clearpage
\begin{figure*}[h]
    \centering
    \includegraphics[clip, width=12cm]{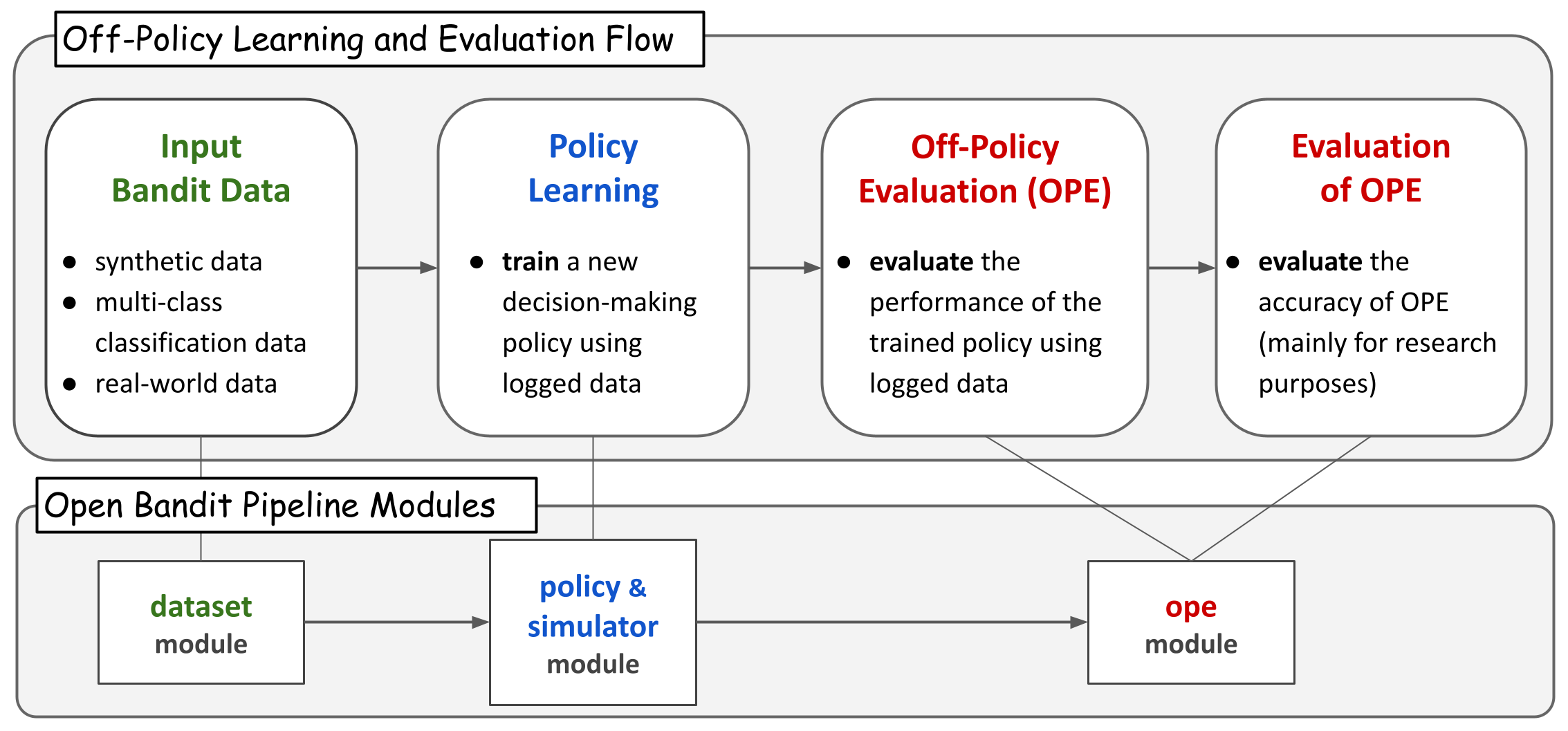}
    \caption{Open Bandit Pipeline Overview.}
    \label{fig:overview_obp}
\end{figure*}

\section{Open Bandit Pipeline (OBP) Package} \label{app:obp_usage}

As described in Section~\ref{sec:dataset}, Open Bandit Pipeline contains implementations of dataset preprocessing, several policy learning methods, and a variety of OPE estimators including several advanced methods.

\subsection{Core Functions}

\paragraph{Modules}
Open Bandit Pipeline consists of the following three main modules.

\begin{itemize}
    \item \textbf{dataset module}\footnote{\url{https://github.com/st-tech/zr-obp/tree/master/obp/dataset}}: This module provides a data loader for Open Bandit Dataset and a flexible interface for handling logged bandit data. It also provides tools to generate synthetic bandit data and transform multiclass classification data to bandit data.
    \item \textbf{policy module}\footnote{\url{https://github.com/st-tech/zr-obp/tree/master/obp/policy}}: This module provides interfaces for implementing new online and offline bandit policies. It also implements several standard policy learning methods.
    \item \textbf{ope module}\footnote{\url{https://github.com/st-tech/zr-obp/tree/master/obp/ope}}: This module provides interfaces for implementing OPE estimators. It also implements several standard and advanced OPE estimators.
\end{itemize}

\paragraph{Supported Estimators and Algorithms}

Open Bandit Pipeline implements the following bandit algorithms and OPE estimators.

\textbf{Bandit Algorithms}
\begin{itemize}
    \item Online Bandit Algorithms
        \begin{itemize}
            \item Non-contextual
                \begin{itemize}
                    \item Uniform Random
                    \item Epsilon Greedy
                    \item Bernoulli Thompson Sampling
                \end{itemize}
            \item Contextual
                \begin{itemize}
                    \item Linear/Logistic Epsilon Greedy
                    \item Linear/Logistic Thompson Sampling
                    \item Linear/Logistic Upper Confidence Bound
                \end{itemize}
        \end{itemize}
    \item Offline (Batch) Bandit Algorithms
        \item IPW Learner (used with a \textit{scikit-learn} classifier)
        \item Neural Network Policy Learner (implemented with \textit{PyTorch})
\end{itemize}

\textbf{OPE Estimators}
\begin{itemize}
    \item Standard OPE (formulated in Section~\ref{sec:ope})
        \begin{itemize}
            \item Direct Method
            \item Inverse Probability Weighting~\citep{precup2000eligibility}
            \item Self-Normalized Inverse Probability Weighting~\citep{swaminathan2015self}
            \item Doubly Robust~\citep{dudik2014doubly}
            \item Self-Normalized Doubly Robust
            \item Switch Doubly Robust~\citep{wang2016optimal}
            \item Doubly Robust with Optimistic Shrinkage~\citep{su2020doubly}
        \end{itemize}
    \item OPE for Slate Recommendation
        \begin{itemize}
            \item Independent Inverse Propensity Score~\citep{li2018offline}
            \item Reward Interaction Inverse Propensity Score~\citep{mcinerney2020counterfactual}
        \end{itemize}
    \item OPE for Continuous Action
        \begin{itemize}
            \item Kernelized Inverse Probability Weighting~\citep{kallus2018policy}
            \item Kernelized Self-Normalized Inverse Probability Weighting~\citep{kallus2018policy}
            \item Kernelized Doubly Robust~\citep{kallus2018policy}
        \end{itemize}
\end{itemize}

Please refer to the package documentation for the basic formulation of OPE and the definitions of the estimators.\footnote{https://zr-obp.readthedocs.io/en/latest/} Note that, in addition to the above algorithms and estimators, the pipeline also provides flexible interfaces. Therefore, researchers can easily implement their own algorithms or estimators and evaluate them with our data and pipeline. Moreover, the pipeline provides an interface for handling real-world logged bandit data. Thus, practitioners can combine their own data with the pipeline and easily evaluate bandit algorithms' performance in their settings with OPE.

\subsection{Example Code (OPE Experiment with Open Bandit Dataset)}
Below, we show an example of conducting an OPE of the performance of BernoulliTS using IPW as an OPE estimator and the Random policy as a behavior policy. 
We see that only ten lines of code are sufficient to complete the standard OPE procedure from scratch (Code Snippet 1).

\begin{minipage}[t]{\textwidth}
\begin{lstlisting}[title={Code Snippet 1: \textbf{Overall Flow of Off-Policy Evaluation using Open Bandit Pipeline}},captionpos=b]
# implementing OPE of BernoulliTS using log data generated by Random
>>> from obp.dataset import OpenBanditDataset
>>> from obp.policy import BernoulliTS
>>> from obp.ope import OffPolicyEvaluation, InverseProbabilityWeighting as IPW

# (1) Data Loading and Preprocessing
>>> dataset = OpenBanditDataset(behavior_policy="random", campaign="all")
>>> bandit_feedback = dataset.obtain_batch_bandit_feedback()

# (2) Production Policy Replication
>>> evaluation_policy = BernoulliTS(
        n_actions=dataset.n_actions,
        len_list=dataset.len_list,
        is_zozotown_prior=True, # replicate policy used in the ZOZOTOWN production
        campaign="all",
        random_state=0
    )
>>> action_dist = evaluation_policy.compute_batch_action_dist(
        n_sim=100000, n_rounds=bandit_feedback["n_rounds"]
    )

# (3) Off-Policy Evaluation
>>> ope = OffPolicyEvaluation(
        bandit_feedback=bandit_feedback, 
        ope_estimators=[IPW()]
    )
>>> estimated_policy_value = ope.estimate_policy_values(action_dist=action_dist)

# estimate the performance improvement of BernoulliTS over Random
>>> ground_truth_random = bandit_feedback["reward"].mean()
>>> print(estimated_policy_value["ipw"] / ground_truth_random) 
1.198126...
\end{lstlisting}
\end{minipage}

In the following subsections, we explain some important features in the example flow.

\subsubsection{Data Loading and Preprocessing}

We prepare easy-to-use data loader for Open Bandit Dataset.
\texttt{obp.dataset.OpenBanditDataset} will download and preprocess the original data.

\begin{minipage}[t]{\textwidth}
\begin{lstlisting}[title={Code Snippet 2: \textbf{Data Loading and Preprcessing}},captionpos=b]
# load and preprocess raw data in the "ALL" campaign collected by the Random policy
>>> dataset = OpenBanditDataset(behavior_policy="random", campaign="all")
# obtain logged bandit feedback generated by the behavior policy
>>> bandit_feedback = dataset.obtain_batch_bandit_feedback()
\end{lstlisting}
\end{minipage}

Users can implement their own feature engineering in the \texttt{pre\_process} method of \texttt{obp.dataset.OpenBanditDataset}. 
Moreover, by following the interface of \texttt{BaseBanditDataset} in the dataset module, one can handle future open datasets for bandit algorithms and OPE.
The dataset module also provides a class to generate synthetic bandit datasets and to modify multiclass classification data to bandit data.

\subsubsection{Production Policy Replication}

After preparing the logged bandit data, we now replicate BernoulliTS used during the data collection in the ZOZOTOWN production. 
Then, we can use it as the evaluation policy.

\begin{minipage}[t]{\textwidth}
\begin{lstlisting}[title={Code Snippet 3: \textbf{Production Policy Replication}},captionpos=b]
# define evaluation policy (the Bernoulli TS policy here)
>>> evaluation_policy = BernoulliTS(
        n_actions=dataset.n_actions,
        len_list=dataset.len_list,
        is_zozotown_prior=True, # replicate BernoulliTS in the ZOZOTOWN production
        campaign="all",
        random_state=0
    )
# compute action choice probabilities of the evaluation policy by running simulation
# action_dist is an array of shape (n_rounds, n_actions, len_list)
# representing action choice probabilities of the evaluation policy
>>> action_dist = evaluation_policy.compute_batch_action_dist(
        n_sim=100000, n_rounds=bandit_feedback["n_rounds"]
    )
\end{lstlisting}
\end{minipage}

The \texttt{compute\_batch\_action\_dist} method of BernoulliTS computes action choice probabilities based on given hyperparameters of the beta distribution.
By activating the \texttt{is\_zozotown\_prior} argument, one can replicate BernoulliTS used in the ZOZOTOWN production.
\texttt{action\_dist} is an array representing the distribution over actions (i.e., action choice probabilities) made by the evaluation policy.

\subsubsection{Off-Policy Evaluation}

Our final step is OPE, which attempts to estimate the performance of bandit policies using only the log data generated by a behavior policy.
Our pipeline provides an easy procedure to implement OPE as follows.

\begin{minipage}[t]{\textwidth}
\begin{lstlisting}[title={Code Snippet 4: \textbf{Off-Policy Evaluation}},captionpos=b]
# estimate the policy value of BernoulliTS based on its action choice probabilities
# it is possible to set multiple OPE estimators to the `ope_estimators` argument
>>> ope = OffPolicyEvaluation(
        bandit_feedback=bandit_feedback, 
        ope_estimators=[IPW()]
    )
>>> estimated_policy_value = ope.estimate_policy_values(action_dist=action_dist)
>>> print(estimated_policy_value)
{"ipw": 0.004553...} # dictionary containing policy values estimated by each estimator

# compare the estimated performance of BernoulliTS with the performance of Random
# our OPE procedure suggests that BernoulliTS improves Random by 19.81%
>>> ground_truth_random = bandit_feedback["reward"].mean()
>>> print(estimated_policy_value["ipw"] / ground_truth_random)
1.198126...
\end{lstlisting}
\end{minipage}

Users can implement their own OPE estimator by following the interface of \texttt{BaseOffPolicyEstimator} class.
\texttt{OffPolicyEvaluation} class summarizes and compares the policy values estimated by several OPE estimators. 
\texttt{bandit\_feedback["reward"].mean()} is the empirical mean of factual rewards (on-policy estimate of the policy value) in the log and thus is the performance of the Random policy during the data collection period.

\subsection{Example Code (Bandit Experiment with Open Bandit Dataset)}
By using Open Bandit Dataset, we can evaluate the performance of (online and offline) bandit algorithms.
Here is an example of using Open Bandit Dataset to evaluate the performance of IPWLearner, which is an offline bandit learning method described in Appendix~\ref{app:examples}.

\begin{minipage}[t]{\textwidth}
\begin{lstlisting}[title={Code Snippet 5: \textbf{Bandit Experiment with Open Bandit Dataset}},captionpos=b]
# implementing the evaluation of IPWLearner using log data generated by Random
>>> from sklearn.linear_model import LogisticRegression
# import open bandit pipeline (obp)
>>> from obp.dataset import OpenBanditDataset
>>> from obp.policy import IPWLearner
>>> from obp.ope import OffPolicyEvaluation, InverseProbabilityWeighting as IPW

# (1) Data Loading and Preprocessing
>>> dataset = OpenBanditDataset(behavior_policy="random", campaign="all")
>>> bandit_feedback_train, bandit_feedback_test = dataset.obtain_batch_bandit_feedback(is_timeseries_split=True)

# (2) Off-Policy Learning
>>> eval_policy = IPWLearner(
        n_actions=dataset.n_actions, base_classifier=LogisticRegression()
    )
>>> eval_policy.fit(
        context=bandit_feedback_train["context"],
        action=bandit_feedback_train["action"],
        reward=bandit_feedback_train["reward"],
        pscore=bandit_feedback_train["pscore"]
    )
>>> action_dist = eval_policy.predict(context=bandit_feedback_test["context"])

# (3) Off-Policy Evaluation
>>> ope = OffPolicyEvaluation(
        bandit_feedback=bandit_feedback, 
        ope_estimators=[IPW()]
    )
>>> estimated_policy_value = ope.estimate_policy_values(action_dist=action_dist)

# performance improvement of IPWLearner over Random (baseline)
>>> ground_truth_random = bandit_feedback["reward"].mean()
>>> print(estimated_policy_value["ipw"] / ground_truth_random) 
1.198126...
\end{lstlisting}
\end{minipage}

By following the above flow, researchers can evaluate the performance of their (online or offline) bandit algorithms with Open Bandit Dataset and Pipeline.

A formal quickstart example with synthetic classification data is available at \url{https://github.com/st-tech/zr-obp/blob/master/examples/quickstart/opl.ipynb}. We also prepare a script to conduct the evaluation of bandit algorithms with synthetic data in \url{https://github.com/st-tech/zr-obp/tree/master/examples/opl} (evaluation of offline bandit algorithms) and \url{https://github.com/st-tech/zr-obp/tree/master/examples/online} (evaluation of online bandit algorithms).

\subsection{Example Code (OPE Experiment with Synthetic Data)}
With Open Bandit Pipeline, we can use synthetic data to evaluate the performance of OPE estimators. Here is an example of conducting OPE of the performance of IPWLearner using Inverse Probability Weighting (IPW) as an OPE estimator. We then evaluate the OPE performance of IPW based on the squared error.

\begin{minipage}[t]{\textwidth}
\begin{lstlisting}[title={Code Snippet 6: \textbf{OPE Experiment with Synthetic Data}},captionpos=b]
# implementing OPE of IPWLearner using synthetic bandit data
>>> from sklearn.linear_model import LogisticRegression
# import open bandit pipeline (obp)
>>> from obp.dataset import SyntheticBanditDataset
>>> from obp.policy import IPWLearner
>>> from obp.ope import (
        OffPolicyEvaluation,
        RegressionModel,
        InverseProbabilityWeighting as IPW,
    )

# (1) Generate Synthetic Bandit Data
dataset = SyntheticBanditDataset(n_actions=10, reward_type="binary")
>>> bandit_feedback_train = dataset.obtain_batch_bandit_feedback(n_rounds=1000)
>>> bandit_feedback_test = dataset.obtain_batch_bandit_feedback(n_rounds=1000)

# (2) Off-Policy Learning
>>> eval_policy = IPWLearner(
        n_actions=dataset.n_actions, base_classifier=LogisticRegression()
    )
>>> eval_policy.fit(
        context=bandit_feedback_train["context"],
        action=bandit_feedback_train["action"],
        reward=bandit_feedback_train["reward"],
        pscore=bandit_feedback_train["pscore"]
    )
>>> action_dist = eval_policy.predict(context=bandit_feedback_test["context"])

# (3) Off-Policy Evaluation and Evaluation of OPE
>>> ope = OffPolicyEvaluation(
        bandit_feedback=bandit_feedback, 
        ope_estimators=[IPW()]
    )
# evaluate the estimation performance (accuracy) of IPW by the squared error ("se")
>>> squared_errors = ope.evaluate_performance_of_estimators(
            ground_truth_policy_value=ground_truth,
            action_dist=action_dist,
            metric="se",
    )
>>> print(squared_errors)
{'ipw': 0.00267235896316153} # the accuracy of IPW in OPE
\end{lstlisting}
\end{minipage}

By following the above flow, researchers can evaluate the performance of their OPE estimators on synthetic data.
A formal quickstart example with synthetic classification data is available at \url{https://github.com/st-tech/zr-obp/blob/master/examples/quickstart/synthetic.ipynb}. 
We also prepare a script to conduct the evaluation of OPE experiments with synthetic data in \url{https://github.com/st-tech/zr-obp/tree/master/examples/synthetic}.

\subsection{Example Code (OPE Experiment with Multiclass Classification Data)}
Researchers often use multiclass classification data to evaluate the estimation accuracy of OPE estimators~\citep{agarwal2017effective,dudik2014doubly,farajtabar2018more,kallus2019intrinsically,su2020doubly,su2019cab,wang2016optimal}. 
Appendix G of Farajtabar et al.~\citep{farajtabar2018more} describes how to modify classification datasets into bandit data in detail.
Open Bandit Pipeline facilitates this kind of OPE experiments with multiclass classification data as follows.

\begin{minipage}[t]{\textwidth}
\begin{lstlisting}[title={Code Snippet 7: \textbf{OPE Experiment with Classification Data}},captionpos=b]
# implementing evaluation of OPE using classification data
>>> from sklearn.datasets import load_digits
>>> from sklearn.ensemble import RandomForestClassifier
>>> from sklearn.linear_model import LogisticRegression
# import open bandit pipeline (obp)
>>> from obp.dataset import MultiClassToBanditReduction
>>> from obp.ope import OffPolicyEvaluation, InverseProbabilityWeighting as IPW


# (1) Data Loading and Bandit Reduction
>>> X, y = load_digits(return_X_y=True)
>>> dataset = MultiClassToBanditReduction(
        X=X, 
        y=y, 
        base_classifier_b=LogisticRegression(random_state=0)
    )
>>> dataset.split_train_eval(eval_size=0.7, random_state=0)
>>> bandit_feedback = dataset.obtain_batch_bandit_feedback(random_state=0)

# (2) Evaluation Policy Derivation
# obtain action choice probabilities of an evaluation policy
>>> action_dist = dataset.obtain_action_dist_by_eval_policy(
        base_classifier_e=RandomForestClassifier(random_state=0)
    )
# calculate the ground-truth performance of the evaluation policy
>>> ground_truth = dataset.calc_ground_truth_policy_value(action_dist=action_dist)
>>> print(ground_truth)
0.9634340222575517

# (3) Off-Policy Evaluation and Evaluation of OPE
>>> ope = OffPolicyEvaluation(
        bandit_feedback=bandit_feedback, 
        ope_estimators=[IPW()]
    )
# evaluate the estimation performance (accuracy) of IPW by the squared error ("se")
>>> squared_errors = ope.evaluate_performance_of_estimators(
            ground_truth_policy_value=ground_truth,
            action_dist=action_dist,
            metric="se",
    )
>>> print(squared_errors)
{'ipw': 0.01827255896321327} # the accuracy of IPW in OPE
\end{lstlisting}
\end{minipage}

By following the above flow, researchers can evaluate the performance of their OPE estimators by transforming classification data into bandit data.

A formal quickstart example with multiclass classification data is available at \url{https://github.com/st-tech/zr-obp/blob/master/examples/quickstart/multiclass.ipynb}. We also prepare a script to conduct the evaluation of OPE experiments with multiclass classification data in \url{https://github.com/st-tech/zr-obp/tree/master/examples/multiclass}.

\end{document}